\documentclass{article}

\PassOptionsToPackage{numbers, compress}{natbib}

\usepackage[final]{neurips_2024}

\usepackage[utf8]{inputenc} 
\usepackage[T1]{fontenc}    
\usepackage{hyperref}       
\usepackage{url}            
\usepackage{booktabs}       
\usepackage{amsfonts}       
\usepackage{amsmath}       %
\usepackage{nicefrac}       
\usepackage{microtype}      
\usepackage{xcolor}         
\usepackage{gensymb}
\usepackage{bm}
\usepackage{amssymb}
\usepackage{subcaption}
\usepackage{graphicx}
\usepackage{multirow}

\newcommand{\figref}[1]{Figure~\ref{fig:#1}}
\newcommand{\tabref}[1]{Table~\ref{tab:#1}} 
\newcommand{\eqnref}[1]{Eq.~\ref{eq:#1}}

\definecolor{purp}{rgb}{0.95, 0.16, 0.65}

\definecolor{tabzeroth}{rgb}{1, 0.5, 0.5}
\definecolor{tabfirst}{rgb}{1, 0.7, 0.7}
\definecolor{tabsecond}{rgb}{1, 0.85, 0.7}
\definecolor{tabthird}{rgb}{1, 1, 0.7} 

\title{ODGS: 3D Scene Reconstruction from Omnidirectional Images with 3D Gaussian Splatting}

\author{%
  Suyoung Lee\thanks{indicates equal contribution.}~~$^1$ \qquad Jaeyoung Chung\footnotemark[1]~~$^1$ \qquad Jaeyoo Huh~$^2$ \qquad Kyoung Mu Lee~$^{1,2}$ \\
  $^1$Dept. of ECE \& ASRI, $^2$IPAI, Seoul National University, Seoul, Korea\\
  \texttt{\{esw0116,\,robot0321\}@snu.ac.kr}\qquad \texttt{jaeyoo900@gmail.com} \qquad \texttt{kyoungmu@snu.ac.kr} \\
}

\begin{document}

\maketitle

\begin{abstract}
\setcounter{footnote}{0} 
Omnidirectional (or 360-degree) images are increasingly being used for 3D applications since they allow the rendering of an entire scene with a single image.
Existing works based on neural radiance fields demonstrate successful 3D reconstruction quality on egocentric videos, yet they suffer from long training and rendering times.
Recently, 3D Gaussian splatting has gained attention for its fast optimization and real-time rendering.
However, directly using a perspective rasterizer to omnidirectional images results in severe distortion due to the different optical properties between two image domains.
In this work, we present ODGS, a novel rasterization pipeline for omnidirectional images, with geometric interpretation.
For each Gaussian, we define a tangent plane that touches the unit sphere and is perpendicular to the ray headed toward the Gaussian center.
We then leverage a perspective camera rasterizer to project the Gaussian onto the corresponding tangent plane.
The projected Gaussians are transformed and combined into the omnidirectional image, finalizing the omnidirectional rasterization process.
This interpretation reveals the implicit assumptions within the proposed pipeline, which we verify through mathematical proofs.
The entire rasterization process is parallelized using CUDA, achieving optimization and rendering speeds 100 times faster than NeRF-based methods.
Our comprehensive experiments highlight the superiority of ODGS by delivering the best reconstruction and perceptual quality across various datasets.
Additionally, results on roaming datasets demonstrate that ODGS restores fine details effectively, even when reconstructing large 3D scenes.
The source code is available on our project page.\footnote{\url{https://github.com/esw0116/ODGS}}

\end{abstract}
\section{Introduction}

With the development of VR/MR devices, robotics technologies, and increasing demands of such applications, 3D scene reconstruction has become one of the crucial tasks in computer vision.
Traditional works have employed a structure-from-motion algorithm that estimates camera motion and scene geometry from multiview 2D images by finding the correspondences between images.
As target 3D scenes become broader and more complex, accurate reconstruction demands a larger volume of images and increases the computational burden required for identifying correspondences.
Recently, some approaches have tried to alleviate these challenges by utilizing wide-angle cameras to capture wide field-of-view images.
Omnidirectional images, which provide a 360-degree field of view, are gaining increased interest because they encompass whole scenes within a single image, thereby reducing the cost of inter-image feature matching.
The growing popularity of 360-degree cameras for personal video recording and the concurrent release of related datasets further facilitate the research on 3D content reconstruction from omnidirectional images.

Such 3D reconstruction techniques~\cite{moulon2016openmvg,wu2023360} began to be mainly studied in SLAM systems to obtain accurate camera poses with matched 3D points from monocular omnidirectional video obtained from robots.
However, these models focus on restoring structural information rather than contents, and they often bypass the fine details and texture of 3D scenes.
After neural radiance field (NeRF)~\cite{mildenhall2020nerf} has shown outstanding 3D reconstruction performance, several works such as \cite{choi2023balanced,hsu2021moving,huang2022tc360roam,kulkarni2023360fusionnerf} attempted to reconstruct 3D implicit representation from omnidirectional images.
Despite showing prominent reconstruction quality, those methods commonly suffer from slow rendering and lengthy training.
3D Gaussian splatting~\cite{kerbl20233d}, 3DGS in short, overcomes the challenges of NeRF by representing 3D contents with numerous Gaussian splats.
The 3D Gaussians are initialized from the sparse point cloud obtained from the structure-from-motion and optimized through the differentiable image rasterization pipeline.
Since the CUDA-implemented rasterization for 3DGS is much faster than volume rendering used in the NeRF family, 3DGS has dramatically improved rendering speed while maintaining or improving performance.
Although many follow-up works have been proposed after 3DGS's success, only a few address 3DGS in the omnidirectional image domain.

In this work, we propose ODGS that aims to reconstruct high-quality 3D scenes represented by Gaussian splatting from multiple omnidirectional images.
The gist of our method is designing a CUDA rasterizer that is appropriate for omnidirectional images.
Specifically, we create a unit sphere from the camera origin, considering it as an omnidirectional camera surface, and assume that each Gaussian is projected onto the tangent plane of the point where the vector from the camera origin to the center of Gaussian and the unit sphere meet.
Since each Gaussian is projected onto a different plane, we calculate a rotation matrix for the coordinate transformation to ensure that each Gaussian is properly projected onto its corresponding tangent plane.
Then the projected Gaussians are subsequently mapped onto the omnidirectional image plane.
Our proposed rasterizer is also easily parallelizable like the original 3DGS rasterizer, demonstrating fast optimization and rendering speed.
Finally, we carefully apply the densification rule to split or prune the Gaussians for omnidirectional projection.
We apply a dynamic gradient threshold value for each Gaussian based on its elevation, as the azimuthal width of the projected Gaussian is stretched when transformed into an equirectangular space.
We conduct comprehensive experiments comparing the reconstruction quality in various 360-degree video datasets with various environments, including egocentric and roaming, real and synthetic.
The results show that ODGS achieves much faster optimization speed than existing NeRF-based methods and reconstructs the scenes with higher accuracy.
Additionally, the perceptual metrics and qualitative results demonstrate that our method restores textural details more sharply.

To summarize, our contributions are three-fold:
\begin{itemize}
\item We introduce ODGS, a 3D reconstruction framework for omnidirectional images based on 3D Gaussian splatting, achieving 100 times faster optimization and rendering speed than NeRF-based methods.
\item We present a detailed geometric interpretation of the rasterization for omnidirectional images, along with mathematical verification, and propose a CUDA rasterizer based on the interpretation.
\item We comprehensively validate ODGS on various egocentric and roaming datasets, showing both more accurate reconstructed results and better perceptual quality.
\end{itemize}

\section{Related works}

In computer vision, ongoing research has been on creating 3D representations of the surrounding environment using multi-view images.
Among them, omnidirectional images capture the surrounding space in a single image due to their wide field of view, making them increasingly popular for 3D reconstruction and mapping.
Traditional structure-from-motion (SfM) algorithms~\cite{pagani2011structure, scaramuzza2006flexible,  schonberger2016structure} simultaneously estimate camera poses and 3D geometry structure by extracting and matching feature points across multiple images. 
This field has developed over many years, resulting in the release of user-friendly open libraries such as COLMAP~\cite{schonberger2016structure} or OpenMVG~\cite{moulon2016openmvg}. 
Recent advancements continue to improve feature matching for spherical images~\cite{gava2023sphereglue}.
In indoor environments, additional information such as room layout~\cite{bai2024360, pintore2021deep3dlayout, rao2021omnilayout} and planar surfaces~\cite{eder2019pano, sun2021indoor} are used to promote the reconstruction quality. 
The geometry structures estimated from omnidirectional images are also utilized for localization~\cite{huang2023360loc, kim2021piccolo} or for simultaneous localization and mapping (SLAM) research~\cite{caruso2015large, sumikura2019openvslam, won2020omnislam}. 
The wide field of view provided by omnidirectional cameras enables the simultaneous capture of extensive spatial information, making them highly beneficial in robotic applications for environmental perception and understanding.
Beyond sparse geometry structure in SfM, Multi-View Stereo (MVS)~\cite{furukawa2015multi}  supports dense reconstruction based on epipolar geometry to achieve better results.
Recently, multi-view stereo techniques leveraging deep neural networks have been actively researched.~\cite{chiu2023360mvsnet, li2022mode, meuleman2021real} 
Another approach to representing 3D is by stacking multiple layers of multi-sphere images. 
Inspired by multi-planar images, this method facilitates the egocentric representation of scenes~\cite{attal2020matryodshka, habtegebrial2022somsi}.
These methods show the possibilities of 3D reconstructions using omnidirectional images, but often lack textural details for photo-realistic 3D reconstruction or limit the representation to confined spaces.

In recent 3D reconstruction research, Neural Radiance Field (NeRF)~\cite{mildenhall2020nerf} has demonstrated the capability for photo-realistic novel-view synthesis, leading to studies on NeRF-based 360 image 3D reconstruction.
This approach is widely used for directly reconstructing scenes in 3D~\cite{choi2023balanced, gu2022omni, huang2022tc360roam, li2022omnivoxel} or indirectly representing 3D by estimating depth~\cite{chang2023depth, chen2022casual, kulkarni2023360fusionnerf}.
In particular, EgoNeRF~\cite{choi2023balanced} is a recently published NeRF-based reconstruction method, pointing out that a typical Cartesian coordinate is not appropriate for representing a large scene with omnidirectional images.
It introduces a new spherically balanced feature grid and hierarchical density adaptation during ray casting, achieving a prominent reconstruction quality.
However, although NeRF-based methods have shown more realistic reconstruction than traditional techniques, they have the inherent limitation of requiring extensive time for reconstruction and rendering.

3D Gaussian splatting (3DGS)~\cite{kerbl20233d} is a novel 3D representation that demonstrates photo-realistic novel view synthesis while supporting fast optimization and real-time rendering.
3DGS explicitly expresses a space using a set of Gaussian primitives and quickly creates novel views through a rasterization pipeline without the time-consuming ray-casting process in NeRF.
Due to its high applicability, extensive research is rapidly advancing, covering not only typical reconstruction but also sparse reconstruction~\cite{chung2023depth,zhu2023fsgs}, dynamic scene reconstruction~\cite{huang2023sc,wu20234d,yang2023deformable}, SLAM~\cite{keetha2023splatam,matsuki2023gaussian} and even generation~\cite{chung2023luciddreamer,tang2023dreamgaussian,zhou2024dreamscene360}. 
However, 3D scene reconstruction based on omnidirectional images has been barely studied.
This is partly because developing a suitable rasterizer for omnidirectional images that allows real-time rendering is challenging, and such implementations are not publicly available.
360-GS~\cite{bai2024360} is the first method that proposes omnidirectional reconstruction with 3DGS, employing a two-step strategy.
However, it relies on layout-guided error correction, which limits its applicability to indoor scenes.
In this paper, we present a carefully implemented CUDA rasterizer that rotates the projection plane on a unit sphere, which can efficiently optimize 3D Gaussians without any constraints or assumptions on scenes.
Further, we propose a dynamic densification rule designed for a 360-degree camera derived from our analysis, enabling us to reconstruct high-quality scenes rapidly.
We note that a few concurrent works, such as Gaussian splatting with optimal projection strategy~\cite{huang2024error} or OmniGS~\cite{li2024omnigs}, partly share the contributions with ours.

\section{Methods}

\subsection{Preliminary: Rasterization Process in Typical 3D Gaussian Splatting}
3D Gaussian splatting (3DGS)~\cite{kerbl20233d} is a recently proposed 3D representation that models scenes using a set of 3D anisotropic Gaussians derived from multi-view images. 
It initializes the 3D Gaussians using a traditional structure-from-motion library and optimizes their properties—such as position, color, scale, rotation, and opacity—through photometric loss. 
In this section, we explain the rasterization pipeline for 3D Gaussians in a perspective camera as proposed by 3DGS, followed by a discussion of the differences in the rasterization process for an omnidirectional camera.

A 3D Gaussian is represented by its mean and covariance, where the covariance matrix $\bm{\Sigma}$ is expressed as the product of a rotation matrix $\mathbf{R}$ and a scale matrix $\mathbf{S}$ ($\bm{\Sigma} = \mathbf{R}\mathbf{S}\mathbf{S}^T\mathbf{R}^T$) to facilitate optimization through gradient descent. 
When the 3D Gaussian is projected onto the image plane of a perspective camera, the resulting distribution becomes complex since the perspective projection is not a linear transformation.
Following the approach in EWA splatting~\cite{zwicker2002ewa}, 3DGS approximates the projected distribution on the image plane as a 2D Gaussian. 
While introducing some errors, the local affine approximation simplifies the modeling of the projected 3D Gaussian, ultimately reducing computational complexity and increasing rendering speed.
Based on the perspective camera projection function $\pi(\bm{\mu}) = \mathbf{K}_{1:2}\left[\nicefrac{\mu_x}{\mu_z}, \nicefrac{\mu_y}{\mu_z}, 1\right]^T$ with intrinsic matrix $\mathbf{K}$, the first-order approximation of the projection $\pi$ is given as,
\begin{equation}
    \mathbf{J} = \frac{\partial \pi\left(\bm{\mu}\right)}{\partial \bm{\mu}}
    = \begin{bmatrix}
        \frac{f_x}{\mu_z} & 0 & -\frac{f_x \mu_x}{\mu_z^2} \\
        0 & \frac{f_y}{\mu_z} & -\frac{f_y \mu_y}{\mu_z^2} \\
    \end{bmatrix} \in \mathbb{R}^{2\times3},
    \label{eq:jacob_pers}
\end{equation}
where $f_x, f_y$ are focal lengths of the camera and $\bm{\mu}=\left[\mu_x, \mu_y, \mu_z\right]$ is the mean vector of 3D Gaussian expressed in the camera coordinate system.
As a result, the 2D Gaussian distribution on the image plane is represented with mean $\pi(\bm{\mu}) \in\mathbb{R}^{2}$ and covariance $\mathbf{\Sigma}_{\text{2D}}=\mathbf{J}\mathbf{W}\mathbf{\Sigma}\mathbf{W}^T\mathbf{J}^T \in\mathbb{R}^{2\times2}$, where $\mathbf{W}$ denotes the transformation matrix from world space to camera space.
The 2D Gaussian represents the intensity on the image plane and is normalized as follows to ensure the maximum value at the center becomes 1.
\begin{equation}
    \text{For } \bm{x}\in\mathbb{R}^2, \quad G_\text{2D}(\bm{x}) = \exp\left(-\frac{1}{2}(\bm{x}-\pi(\bm{\mu}))^T \bm{\Sigma}_\text{2D}^{-1} (\bm{x}-\pi(\bm{\mu})) \right).
    \label{eq:gaussian2d}
\end{equation}
This bell-shaped intensity is multiplied by the Gaussian's opacity to determine the pixel-wise opacity $\alpha$.
After frustum culling and sorting by depth, the color of each pixel $C$ is determined as,
\begin{equation}
C = \sum_{j\in\textit{N}} {c_j \alpha_j T_j}, \quad {T_j = \prod_{k=1}^{j-1} (1-\alpha_k)},
\label{eq:alphablending}
\end{equation}
where $c_j$ is the color of each Gaussian.
This accumulation process is performed in the order of depth sorting.
Each pixel is processed independently by a single GPU thread, enabling rapid rendering of 3D Gaussians into images through this rasterization process.
While 3DGS describes the rasterization pipeline of 3D Gaussians for the perspective camera, a rasterization pipeline for an omnidirectional camera requires a distinct approach that regards its different optical characteristics.
The following section explains our carefully designed rasterizer for the omnidirectional images.

\subsection{Designing Rasterizer for Omnidirectional Images}

\begin{figure}[t]
    \centering
    \includegraphics[width=\linewidth]{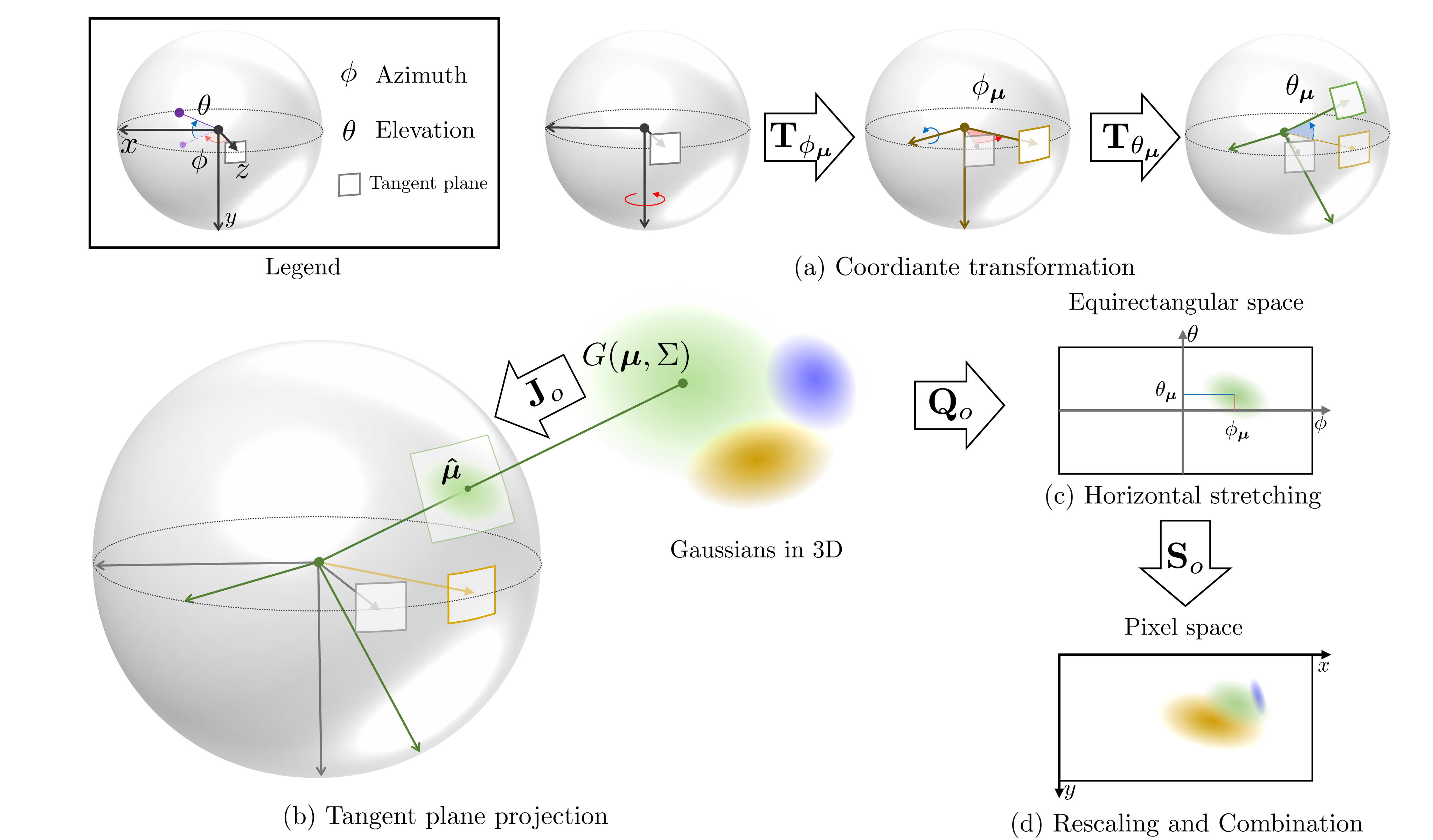}
    \caption{Illustration on rasterization process of ODGS.
    We describe the process of projecting a 3D Gaussian to the omnidirectional pixel space.
    (a) The coordinate is transformed from the original camera pose (black) to the target Gaussian (green), making the $z$-axis of the coordinate heads towards the center of the Gaussian.
    (b) The Gaussian is projected onto the corresponding tangent plane.
    (c) The projected Gaussian is horizontally stretched when transformed into equirectangular space.
    (d) The Gaussian in equirectangular space is linearly transformed to the pixel space, followed by combination with the other projected Gaussian.
    }
    \label{fig:method}
\end{figure}
A 360-degree camera captures all rays from the surrounding 3D environment to the camera origin and represents them on a unit sphere $\mathbb{S}^2$.
We employ spherical projection or equirectangular projection (ERP) to map the projected image on the sphere $\mathbb{S}^2$ to the equirectangular space $\mathbb{R}^2$, then transform to pixel space $\mathbb{R}^2$.
We describe a series of steps on how a 3D Gaussian is approximated as a 2D Gaussian in the pixel space and the feasibility of such an approximation.
The gist of our rasterizer design for the omnidirectional camera is leveraging locally approximated perspective projection on $\mathbb{S}^2$ while minimizing errors.

As demonstrated in \figref{method}, we follow a camera coordinate convention~\cite {bradski2000opencv} with the z-axis forward, the x-axis to the right, and the y-axis down.
We define spherical coordinates by setting the azimuth $\phi$ from the forward z-axis within the range $[-\pi,\pi]$ and the elevation $\theta$ from the z-x plane within the range $[-\pi/2,\pi/2]$.
Projecting the mean $\bm{\mu}$ of the 3D Gaussian onto the unit sphere results in $\hat{\bm{\mu}}=\bm{\mu}/{||\bm{\mu}||}$, which corresponds to the azimuth $\phi_\mu=\arctan{({\mu}_x / {\mu}_z)}$ and elevation $\theta_\mu=\arctan{(-{\mu}_y / \sqrt{{\mu}_x^2+{\mu}_z^2})}$.
This spherical coordinate representation $(\phi, \theta)$ is converted into pixel space by multiplying scalar and adding center shift, 
\begin{equation}
    \pi_o(\bm{\mu}) = \left(\frac{W}{2\pi}\phi_\mu + \frac{W}{2},  -\frac{H}{\pi}\theta_\mu + \frac{H}{2} \right)^T
    \label{eq:equirec_proj}
\end{equation}
where $W, H$ are the width and height of the omnidirectional image, respectively.

While finding the corresponding point of the center of 3D Gaussian on the pixel space is straightforward, calculating the covariance requires more careful consideration.
We model the distribution of 3D Gaussian projected onto pixel space as a 2D Gaussian for computational efficiency and stability, following a similar approach to 3DGS. 
We leverage the perspective camera and local affine approximation to intuitively describe the non-linear transformation introduced by the spherical camera characteristics and the equirectangular projection.
Then, we mathematically prove the correctness of the proposed method.

Let us assume a perspective camera with a unit focal length, where the image plane is tangent to the unit sphere.
We rotate the perspective camera's forward direction to align with the position of the Gaussian center, $\bm{\mu}$, as shown in \figref{method} (a).
The image plane is tangent to the unit sphere at $\hat{\bm{\mu}}$, the point where the line from the sphere's center to the center of 3D Gaussian intersects the sphere.
We define the rotation matrix of the perspective camera as 
$\mathbf{T}_{\bm{\mu}}$, which is accomplished in two rotations in azimuth and elevation,
\begin{align}
\begin{split}
    \mathbf{T}_{\bm{\mu}} = 
    \mathbf{T}_{\theta_\mu} \times \mathbf{T}_{\phi_\mu} &= 
    \begin{bmatrix}
    1 & 0 & 0 \\
    0 & \cos{\theta_\mu} & \sin{\theta_\mu} \\
    0 & -\sin{\theta_\mu} & \cos{\theta_\mu} \\
    \end{bmatrix}
    \times
    \begin{bmatrix}
    \cos{\phi_\mu} & 0 & -\sin{\phi_\mu} \\
    0 & 1 & 0 \\
    \sin{\phi_\mu} & 0 & \cos{\phi_\mu} \\
    \end{bmatrix} \\
    &= \begin{bmatrix}
    \cos{\phi_\mu} & 0 & -\sin{\phi_\mu} \\
    \sin{\theta_\mu}\sin{\phi_\mu} & \cos{\theta_\mu} & \sin{\theta_\mu}\cos{\phi_\mu} \\
    \cos{\theta_\mu}\sin{\phi_\mu} & -\sin{\theta_\mu} & \cos{\theta_\mu}\cos{\phi_\mu} \\
    \end{bmatrix}.
    \label{eq:omni_rotation}
\end{split}
\end{align}

The rotation of the coordinate system helps minimize the error between the unit sphere and the image plane, while also simplifying the covariance calculation.
In the rotated camera coordinate, the position of the Gaussian is represented as $\bm{\mu}_o = \left(0, 0, ||\bm{\mu}|| \right)$.
Thus, the Jacobian matrix from \eqnref{jacob_pers} is simplified as,
\begin{equation}
    \mathbf{J}_o = 
    \begin{bmatrix}
    \nicefrac{1}{||\bm{\mu}||} & 0 & 0 \\
    0 & \nicefrac{1}{||\bm{\mu}||} & 0 \\
    \end{bmatrix},
    \label{eq:omni_J}
\end{equation}
because the focal length of the perspective camera is assumed to be one ($f_x=f_y=1$).
Thus, the covariance of the 3D Gaussian projected onto this tangent plane is modeled as $\mathbf{J}_o\mathbf{W}\mathbf{\Sigma}\mathbf{W}^T\mathbf{J}_o^T$, as shown in \figref{method} (b).
We assume that the covariance of this 2D Gaussian is small enough to disregard the difference between the tangent plane and the sphere surface, allowing us to transfer it directly onto the sphere surface.
Although this assumption does not generally hold, we ensure its validity through the split rule in 3DGS, which keeps the size of the Gaussian small.
Next, we map the 2D covariance from the spherical surface $\mathbb{S}^2$ to the equirectangular space $(\phi,\theta)\in\mathbb{R}^2$, as described in \figref{method} (c). 
The equirectangular projection transforms the spherical surface onto a cylindrical map, scaling a ring at latitude $\theta$ with an initial radius of $\cos{\theta}$ on the sphere to a radius of 1.
This projection introduces a horizontal scaling factor of $\sec{\theta}$, leading to increased distortion as $\theta$ approaches the poles.
We incorporate the distortion through $\mathbf{Q}_o$, and then we rescale the covariance to the pixel space by applying the appropriate scaling factors $\mathbf{S}_o$, 

\begin{equation}
    \mathbf{Q}_o = 
    \begin{bmatrix}
    \sec{\theta_{\mu}} & 0  \\
    0 & 1 \\
    \end{bmatrix},
    \quad
    \mathbf{S}_o = 
    \begin{bmatrix}
    \nicefrac{W}{2\pi} & 0  \\
    0 & \nicefrac{H}{\pi} \\
    \end{bmatrix}.
    \label{eq:omni_QS}
\end{equation}

As a result, the final Jacobian matrix is given as,
\begin{equation}
    \mathbf{J}_{omni} = \mathbf{S}_o \mathbf{Q}_o \mathbf{J}_o \mathbf{T}_{\bm{\mu}} 
    = \begin{bmatrix}
    \frac{W}{2\pi ||\bm{\mu}||}\sec{\theta_\mu}\cos{\phi_\mu} & 0 & -\frac{W}{2\pi ||\bm{\mu}||} \sec{\theta_\mu}\sin{\phi_\mu} \\
    \\
    \frac{H}{\pi||\bm{\mu}||}\sin{\theta_\mu}\sin{\phi_\mu} & \frac{H}{\pi||\bm{\mu}||}\cos{\theta_\mu} & \frac{H}{\pi||\bm{\mu}||}\sin{\theta_\mu}\cos{\phi_\mu} \\
    \end{bmatrix},
    \label{eq:omni_Jacobian}
\end{equation}
where the final 2D covariance is presented as $\mathbf{\Sigma}_{\text{2D},o} = \mathbf{J}_{omni}\mathbf{W}\mathbf{\Sigma}\mathbf{W}^T\mathbf{J}_{omni}^T$
We verify the correctness of the derived method by directly differentiating the equirectangular projection function $\pi_o$ in \eqnref{equirec_proj}, yielding the same result $\mathbf{J}_{omni} = \frac{\partial \pi_{o}\left(\bm{\mu}\right)}{\partial \bm{\mu}}$ as detailed in Appendix~\ref{app:proof}.

As a result of the series of steps, the final 2D covariance is used for rendering the image, as described in \eqnref{gaussian2d} and \eqnref{alphablending}.
One key difference is that, instead of performing frustum-shaped culling as in perspective cameras, we perform culling in a spherical shell.
The rasterization pipeline is fully differentiable and implemented in CUDA which can be used as a typical 3DGS.
The detailed gradient calculations through back-propagation are provided in Appendix~\ref{app:backprop}.

\subsection{Densification Policy for Omnidirectional Images}
\label{sec:densification}

Due to the characteristic of equirectangular projection, a 3D Gaussian can be rendered in different shapes depending on its relative elevation to the camera; Gaussians near the poles are drawn larger.
Therefore, we propose an dynamic densification strategy specifically designed for omnidirectional images.
While the original method uses a pre-defined gradient threshold for densifying Gaussians, we apply a varying gradient threshold $\tau_{\mu}$ according to the elevation angle $\theta_{\mu}$ as,
\begin{equation}
    \tau_{\mu} = \tau_{\text{min}} + (1-\cos{\theta_{\mu}}) \times \left( \tau_{\text{max}} - \tau_{\text{min}} \right),
\end{equation}
which mitigates excessive densification of Gaussians near the poles.

\section{Experiments} 

\subsection{Experiment Details}
\label{sec:expdetails}
\paragraph{Datasets}
We evaluate our method on three egocentric datasets (OmniBlender, Ricoh360, OmniPhotos) and three roaming datasets (360Roam, OmniScenes, 360VO) to show its superiority regardless of domain.
First, EgoNeRF~\cite{choi2023balanced} released OmniBlender and Ricoh360, which have different characteristics.
OmniBlender contains 11 synthetic scenes generated with an omnidirectional rendering engine in Blender~\cite{blender}, with four indoor and seven outdoor scenes.
The images were captured by rotating in a circular motion while ascending, each with a resolution of $2000\times1000$.
Each scene in OmniBlender consists of 25 training and test images.
Ricoh360 contains 12 real-world omnidirectional outdoor scenes captured by rotating in place in a cross-shaped pattern.
Each scene consists of 50 training images and 50 testing images with a resolution of $1920\times960$.
OmniPhotos~\cite{bertel2020omniphotos} has released 10 real-world omnidirectional scenes captured by rotating in a circular motion with a commercial 360-degree camera on a selfie stick.
Each scene has 71 to 91 images with a size of $3840\times1920$.
In our experiment, we resize them to half resolution $1920\times960$, and we use 20\% of images for the test.

For the roaming scenarios, we utilize several multi-view omnidirectional datasets, which were not originally released for 3D reconstruction tasks.
360Roam~\cite{huang2022tc360roam} dataset consists of 10 real-world indoor scenes captured by Insta360camera.
Each scene has 71 to 215 omnidirectional images with size $6080\times3040$, and we resize them to $2048\times1024$. 
OmniScenes~\cite{kim2021piccolo} is originally made for assessing the quality of visual localization of omnidirectional images in harsh conditions.
Since it is proposed to measure the robustness of visual localization algorithms, it contains significant scene changes, motion blur, or some visual artifacts such as jpeg compression.
We use the released version 1.1, which includes 7 real-world indoor captured scenes in resolution $1920\times960$.
360VO~\cite{huang2022360vo} is a simulation dataset for evaluating the localization and mapping algorithms in the robotics field.
It contains 10 virtual outdoor road scenes, where each scene has 2000 images with size $1920\times960$.
We uniformly select 200 images for each sequence for training and testing.
Since these datasets do not split the train and test images, we conducted our experiment by dividing them by 4:1 for train and test, respectively.
We note that all datasets have CC-BY-4.0 licenses.
Although some datasets provide camera poses and dense point clouds, we run the structure-from-motion, specifically OpenMVG~\cite{moulon2016openmvg}, on all the datasets and use obtained poses and point clouds for our experiment.

\begin{table}[t]
    \centering
    \caption{Quantitative comparison of 3D reconstruction methods on various datasets.
    The best metric for each dataset is written in \textbf{bold}.
    Our method shows the best performance on almost all settings regardless of optimization time, with the fastest rendering speed.
    }
    \vspace{+1.0em}
    \resizebox{\textwidth}{!}{
    \begin{tabular}{c|c|ccc|ccc|c}
        \toprule[1.0pt]
        \multirow{2}{*}{Dataset} & \multirow{2}{*}{Methods} & \multicolumn{3}{c|}{10 min} & \multicolumn{3}{c|}{100 min} & Time$_{\downarrow}$ \\
        & & PSNR$_{\uparrow}$ & SSIM$_{\uparrow}$ & LPIPS$_{\downarrow}$ & PSNR$_{\uparrow}$ & SSIM$_{\uparrow}$ & LPIPS$_{\downarrow}$ & (sec.) \\
        \midrule
        \multirow{5}{*}{OmniBlender}
        & NeRF(P) & 19.20 & 0.6124 & 0.5359 & 20.04 & 0.6092 & 0.4949 & 62.71 \\
        & 3DGS(P) & 29.36 & 0.8770 & 0.1400 & 21.19 & 0.7528 & 0.3021 & 0.112 \\
        & TensoRF & 25.36 & 0.7249 & 0.3855 & 26.08 & 0.7416 & 0.3170 & 10.77 \\
        & EgoNeRF & 28.29 & 0.8309 & 0.2194 & 30.89 & 0.8934 & 0.1260 & 23.78 \\
        & ODGS & \textbf{32.76} & \textbf{0.9234} & \textbf{0.0469} & \textbf{33.05} & \textbf{0.9229} & \textbf{0.0343} & \textbf{0.028} \\
        \midrule
        \multirow{5}{*}{Ricoh360}
        & NeRF(P) & 14.33 & 0.5616 & 0.5794 & 16.16 & 0.5617 & 0.5716 & 62.46 \\
        & 3DGS(P) & \textbf{25.12} & 0.7932 & 0.2397 & 22.07 & 0.7228 & 0.3218 & 0.132 \\
        & TensoRF & 23.35 & 0.6812 & 0.5200 & 23.97 & 0.6936 & 0.4653 & 10.30 \\
        & EgoNeRF & 24.74 & 0.7467 & 0.3243 & 25.49 & 0.7737 & 0.2825 & 23.89 \\
        & ODGS & 24.94 & \textbf{0.8135} & \textbf{0.1489} & \textbf{26.27} & \textbf{0.8462} & \textbf{0.1051} & \textbf{0.026} \\
        \midrule
        \multirow{5}{*}{OmniPhotos}
        & NeRF(P) & 18.14 & 0.6158 & 0.5514 & 20.80 & 0.6388 & 0.4772 & 62.08 \\
        & 3DGS(P) & 25.61 & 0.8310 & 0.2100 & 23.30 & 0.7859 & 0.2670 & 0.110 \\
        & TensoRF & 22.78 & 0.6841 & 0.5089 & 23.73 & 0.7038 & 0.4467 & 9.707 \\
        & EgoNeRF & 25.20 & 0.7722 & 0.2662 & 26.90 & 0.8349 & 0.1766 & 23.88 \\
        & ODGS & \textbf{26.24} & \textbf{0.8704} & \textbf{0.1108} & \textbf{27.04} & \textbf{0.8878} & \textbf{0.0875} & \textbf{0.028} \\
        \midrule
        \midrule
        \multirow{5}{*}{360Roam}
        & NeRF(P) & 15.07 & 0.6848 & 0.4839 & 15.26 & 0.6813 & 0.5025 & 62.98 \\
        & 3DGS(P) & 20.17 & 0.7001 & 0.3536 & 19.34 & 0.6576 & 0.3837 & 0.104 \\
        & TensoRF & 18.00 & 0.5988 & 0.7488 & 18.12 & 0.5895 & 0.7133 & 9.052 \\
        & EgoNeRF & 20.45 & 0.6358 & 0.5334 & \textbf{21.18} & 0.6718 & 0.4444 & 24.03 \\
        & ODGS & \textbf{21.08} & \textbf{0.7066} & \textbf{0.3003} & 20.85 & \textbf{0.7111} & \textbf{0.2254} & \textbf{0.029} \\
        \midrule
        \multirow{5}{*}{OmniScenes}
        & NeRF(P) & 15.69 & 0.7218 & 0.4546 & 15.98 & 0.6890 & 0.4914 & 62.90 \\
        & 3DGS(P) & 23.61 & 0.8444 & 0.2835 & 17.14 & 0.7119 & 0.3906 & 0.194 \\
        & TensoRF & 23.58 & 0.8118 & 0.3534 & 24.21 & 0.8208 & 0.3091 & 8.100 \\
        & EgoNeRF & 22.78 & 0.7997 & 0.3463 & \textbf{24.76} & 0.8313 & 0.2623 & 23.66 \\
        & ODGS & \textbf{24.42} & \textbf{0.8526} & \textbf{0.1391} & 24.51 & \textbf{0.8505} & \textbf{0.1282} &  \textbf{0.032} \\
        \midrule
        \multirow{5}{*}{360VO}
        & NeRF(P) & 15.71 & 0.6186 & 0.4949 & 17.78 & 0.6373 & 0.5064 & 61.97 \\
        & 3DGS(P) & 22.87 & 0.7861 & 0.2970 & 22.73 & 0.7822 & 0.3061 & 0.091 \\
        & TensoRF & 19.74 & 0.6543 & 0.5876 & 20.31 & 0.6721 & 0.5640 & 7.815 \\
        & EgoNeRF & 22.47 & 0.7325 & 0.4342 & 23.78 & 0.7677 & 0.3680 & 23.96 \\
        & ODGS & \textbf{24.63} & \textbf{0.8245} & \textbf{0.2175} & \textbf{26.68} & \textbf{0.8694} & \textbf{0.1264} & \textbf{0.026} \\
        \bottomrule
    \end{tabular}
    }
    \label{tab:comp1}
\end{table}

\paragraph{Implementation details}
Our framework is basically built with PyTorch~\cite{paszke2019pytorch}, but we manually implement the omnidirectional rasterizer using the CUDA kernel.
All experiments, including optimization and inference time measurements, are conducted using a single NVIDIA RTX A6000 GPU.
We describe the optimization arguments in the Appendix~\ref{app:parameters}.

\begin{figure}[t]
    \newcommand{\ww}{\linewidth}
    \newcommand{\hh}{0.24\linewidth}
    \centering
    \addtocounter{subfigure}{-3}
    \subfloat{\includegraphics[width=\ww]{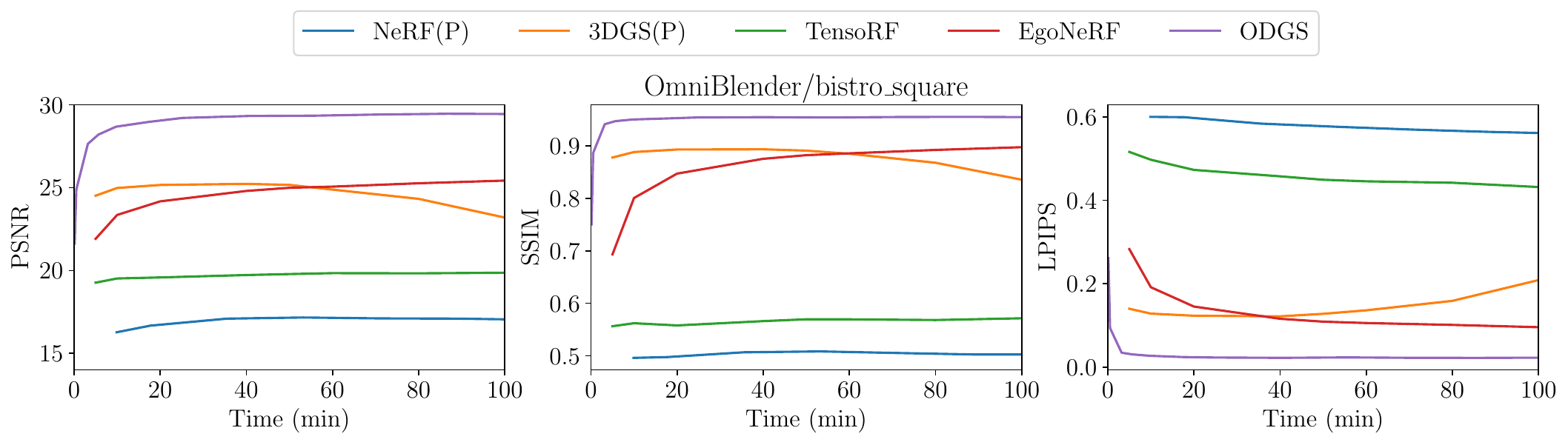}}
    \addtocounter{subfigure}{-3}
    \subfloat{\includegraphics[width=\ww]{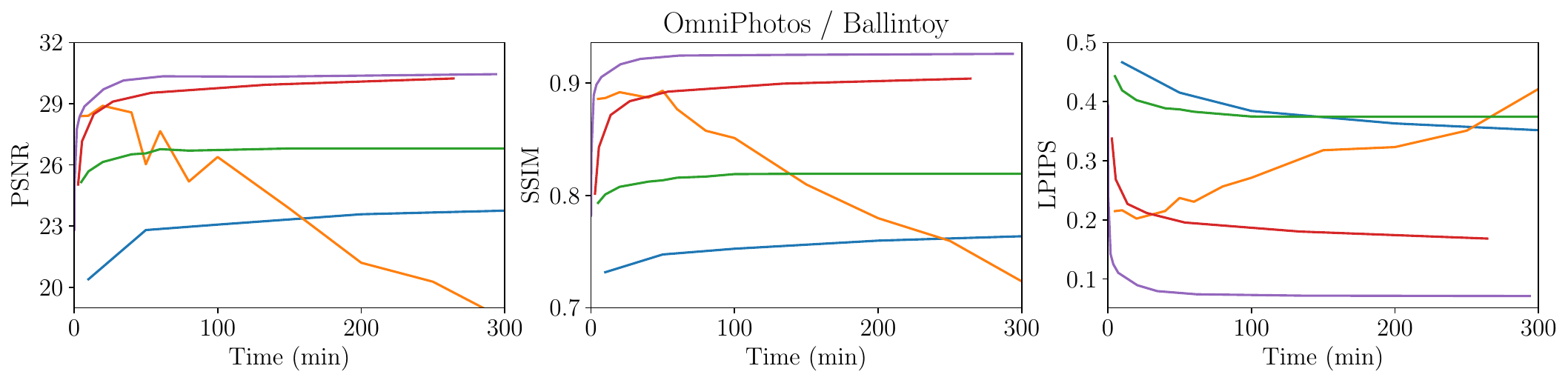}}
    \caption{Changes of PSNR, SSIM, and LPIPS over optimization time for each method.
    ODGS shows the best result as well as the highest convergence speed in both scenes.
    }
    \label{fig:time_metric}
\end{figure}

\subsection{Experiment Results}

\paragraph{Baselines}
With no available code for 3DGS on omnidirectional images at the time of our experiments, we compare our method with NeRF-based methods, specifically TensoRF~\cite{chen2022tensorf} and EgoNeRF~\cite{choi2023balanced}.
We also convert omnidirectional images into perspective images to compare the typical 3D reconstruction methods, NeRF~\cite{mildenhall2020nerf} and 3DGS~\cite{kerbl20233d}.
Specifically, we transform the omnidirectional images into six perspective images using cubemap decomposition, popularly used in many studies involving 360-degree cameras~\cite{jang2022egocentric,wang2020bifuse}.
The six decomposed images compose a cube-shaped surface and we calculate the corresponding camera pose of each surface.
For inference, the six views for each face in the cube are rendered and then combined into an omnidirectional image.

\paragraph{Quantitative comparison}
To ensure the experiment's fairness and highlight the efficiency of our method, all methods were evaluated after optimizing the model with the same amount of time.
We evaluate the performance of all methods at 10 and 100 minutes of training time, measured in wall-clock time.
For evaluation metric, we use PSNR (dB), SSIM~\cite{measure_ssim}, and LPIPS~\cite{feat_deep} for comparing reconstruction quality, where AlexNet~\cite{krizhevsky2012imagenet} backbone is used for measuring LPIPS.
\tabref{comp1} shows the quantitative performance comparison and rendering time (seconds) for the all datasets.
The (P) mark in the method column indicates those methods are trained with converted perspective images.
Our results show dominant results on all metrics, including inference time.
NeRF and TensoRF, which use a grid based on a Cartesian coordinate system, encounter difficulties representing large scenes, resulting in poor quantitative metrics.
EgoNeRF, which introduces a spherical balanced grid to mitigate the challenge, shows better quality than TensoRF but still needs better perceptual metrics.
Also, these methods require more than a second to render a single omnidirectional image for an arbitrary viewpoint, which is impractical for real scenarios.
Meanwhile, 3DGS with perspective images shows the best results except ours when optimized for 10 minutes, but severely suffers from overfitting and gets worse results after 100 minutes of optimization.
In terms of rendering time, despite reporting faster speed than NeRF-based models, original 3DGS takes longer than typical perspective image rendering because it involves non-linear warping of each image when stitching six images to create one omnidirectional image.
ODGS, in contrast, outperforms the other methods in image reconstruction quality and rendering speed.
The outstanding results for SSIM and LPIPS imply that our method generates images with accurate structure and prominent perceptual quality.

\figref{time_metric} shows the change of PSNR, SSIM, and LPIPS depending on the optimization time for two example scenes.
We note that we stopped training NeRF and TensoRF at 100 and 200 minutes, respectively, since their performances converged.
Our method shows the fastest optimization speed in both scenes while maintaining the highest score regardless of optimization time.
Typical NeRF and TensoRF recorded significantly lower results than ours, verifying that the Cartesian coordinate is inappropriate for radially extending rays.

\begin{figure}[!tb]
    \newcommand{\ww}{0.24\linewidth}
    \newcommand{\hh}{0.20\linewidth}
    \centering
    \addtocounter{subfigure}{-4}
    \subfloat{\includegraphics[width=\ww,height=\hh]{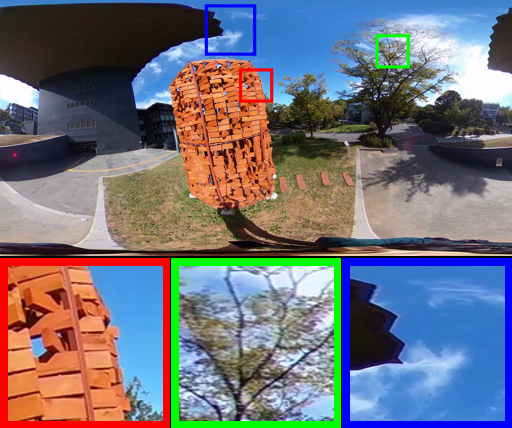}}
    \hfill
    \subfloat{\includegraphics[width=\ww,height=\hh]{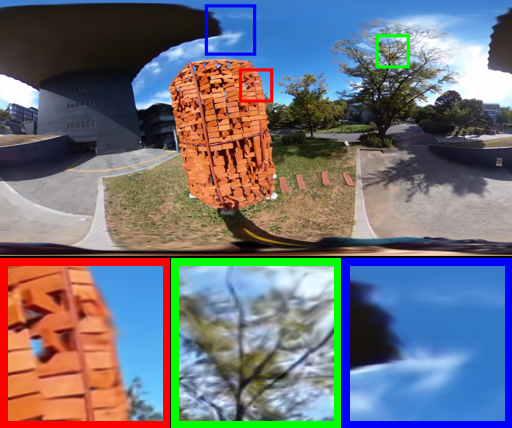}}
    \hfill
    \subfloat{\includegraphics[width=\ww,height=\hh]{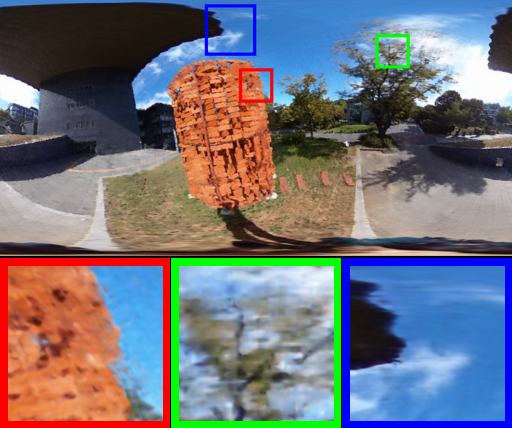}}
    \hfill
    \subfloat{\includegraphics[width=\ww,height=\hh]{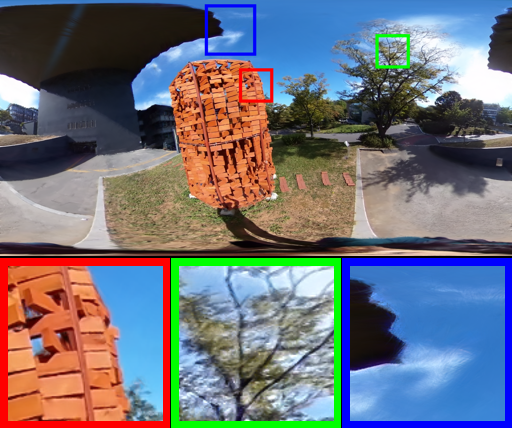}}
    \vspace{1mm}
    \addtocounter{subfigure}{-4}
    \subfloat{\includegraphics[width=\ww,height=\hh]{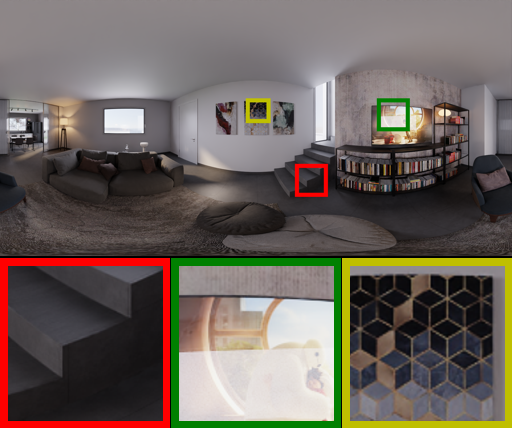}}
    \hfill
    \subfloat{\includegraphics[width=\ww,height=\hh]{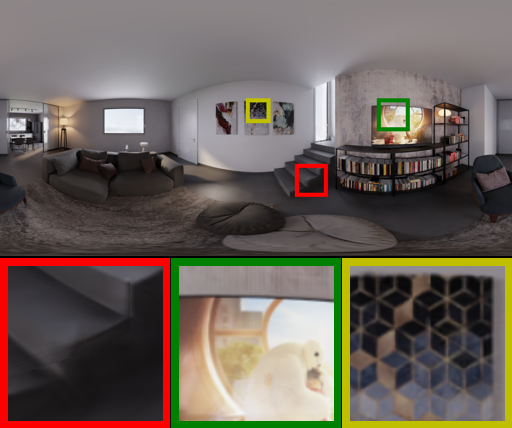}}
    \hfill
    \subfloat{\includegraphics[width=\ww,height=\hh]{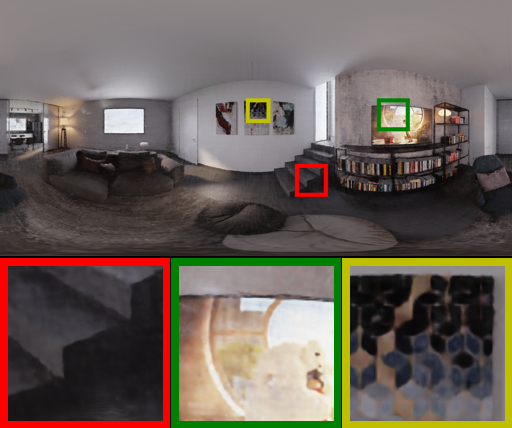}}
    \hfill
    \subfloat{\includegraphics[width=\ww,height=\hh]{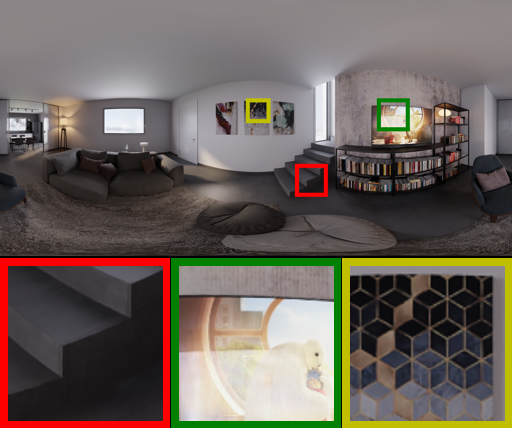}}
    \vspace{1mm}

    \addtocounter{subfigure}{-4}
    \subfloat[Ground truth]{\includegraphics[width=\ww,height=\hh]{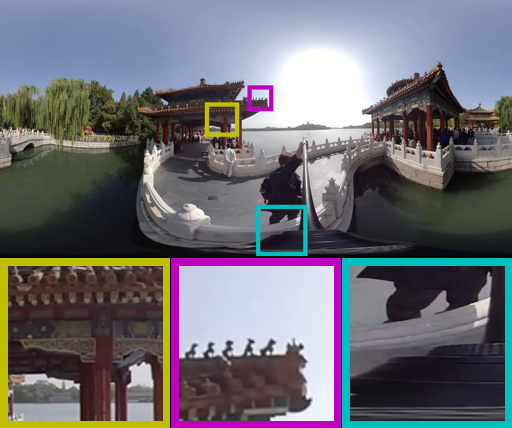}}
    \hfill
    \subfloat[3DGS(P)~\cite{kerbl20233d}]{\includegraphics[width=\ww,height=\hh]{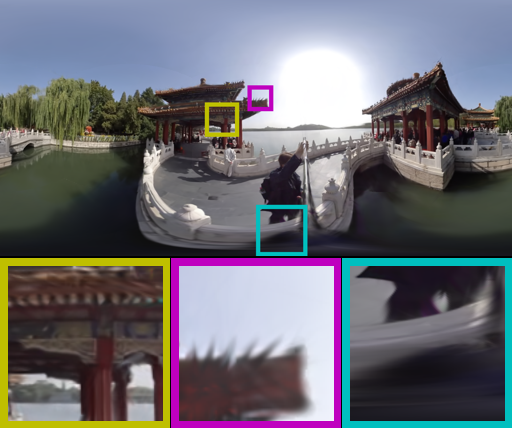}}
    \hfill
    \subfloat[EgoNeRF~\cite{choi2023balanced}]{\includegraphics[width=\ww,height=\hh]{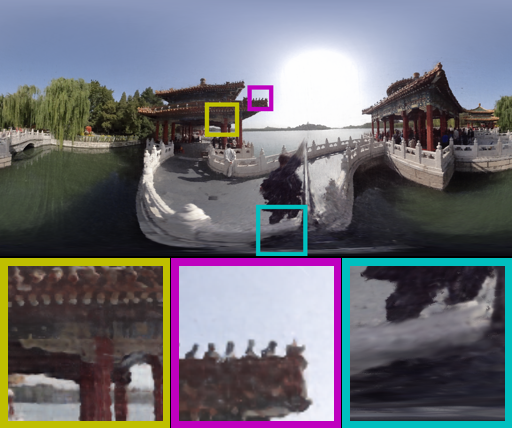}}
    \hfill
    \subfloat[ODGS]{\includegraphics[width=\ww,height=\hh]{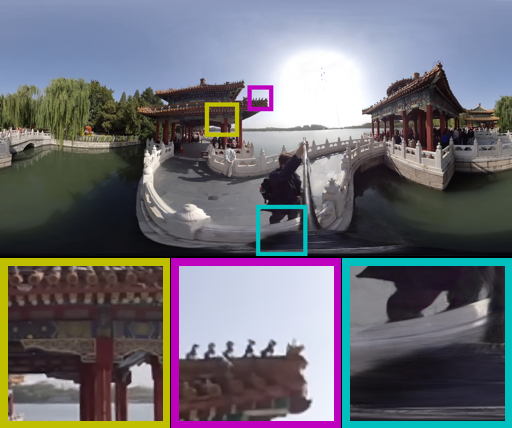}}
    
    \caption{Qualitative comparisons in the egocentric scenes (10 min.).
    Each scene is brought from Ricoh360, OmniBlender, and OmniPhotos, respectively.
    \textit{Best viewed when zoomed in.}
    }
    \label{fig:qual_ego}
    \vspace{-1mm}
\end{figure}
\begin{figure}[!htb]
    \newcommand{\ww}{0.24}
    \newcommand{\hh}{0.20}
    \centering
    \addtocounter{subfigure}{-4}
    \subfloat{\includegraphics[width=\ww\linewidth,height=\hh\linewidth]{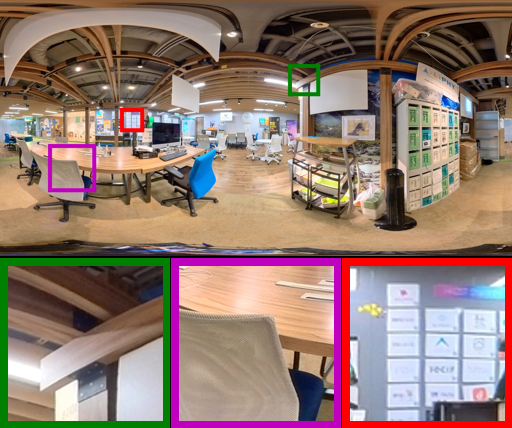}}
    \hfill
    \subfloat{\includegraphics[width=\ww\linewidth,height=\hh\linewidth]{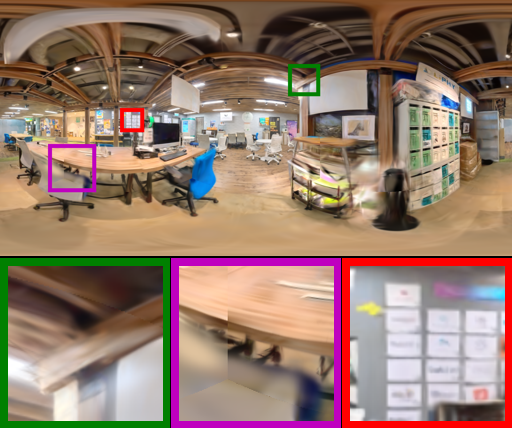}}
    \hfill
    \subfloat{\includegraphics[width=\ww\linewidth,height=\hh\linewidth]{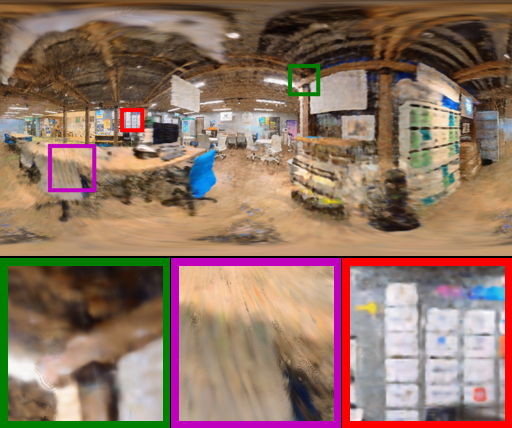}}
    \hfill
    \subfloat{\includegraphics[width=\ww\linewidth,height=\hh\linewidth]{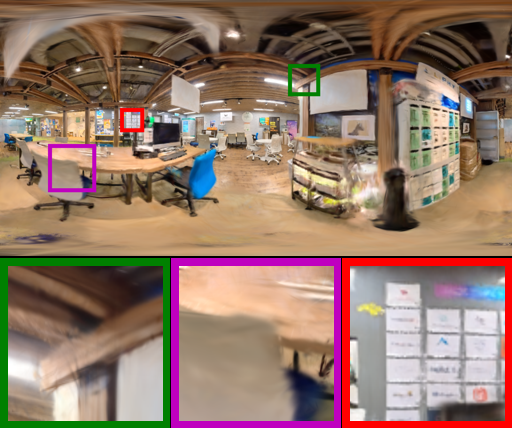}}
    \vspace{1mm}
    \addtocounter{subfigure}{-4}
    \subfloat{\includegraphics[width=\ww\linewidth,height=\hh\linewidth]{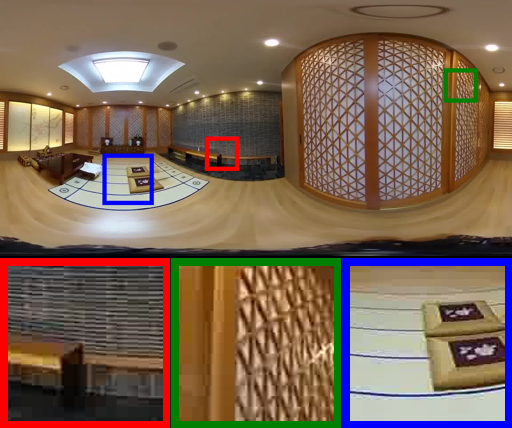}}
    \hfill
    \subfloat{\includegraphics[width=\ww\linewidth,height=\hh\linewidth]{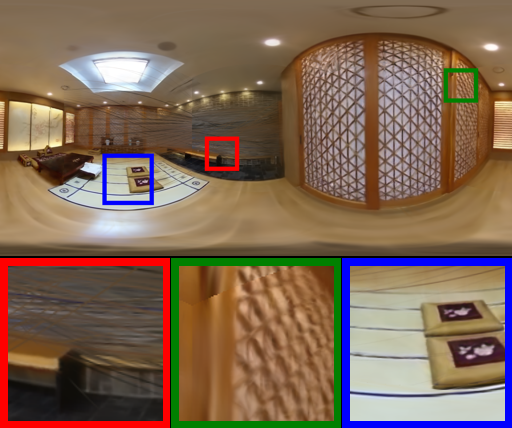}}
    \hfill
    \subfloat{\includegraphics[width=\ww\linewidth,height=\hh\linewidth]{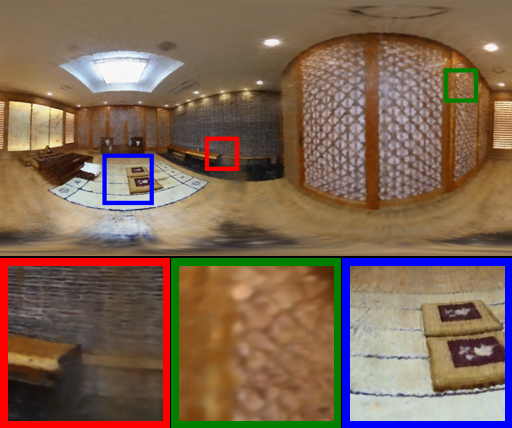}}
    \hfill
    \subfloat{\includegraphics[width=\ww\linewidth,height=\hh\linewidth]{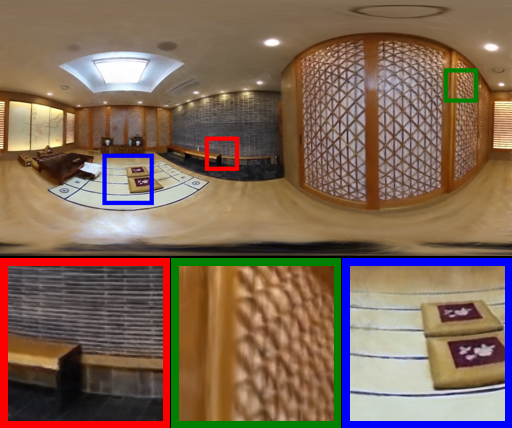}}
    \vspace{1mm}
    \addtocounter{subfigure}{-4}
    \subfloat[Ground truth]{\includegraphics[width=\ww\linewidth,height=\hh\linewidth]{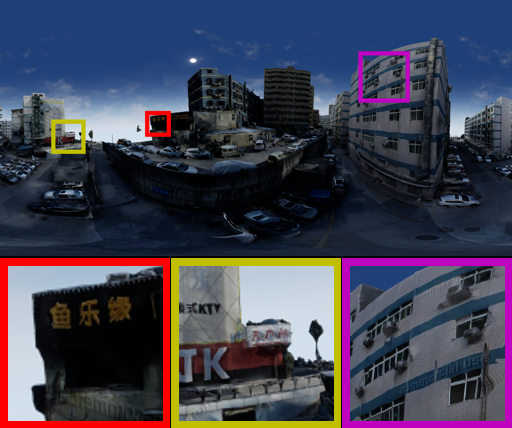}}
    \hfill
    \subfloat[3DGS(P)~\cite{kerbl20233d}]{\includegraphics[width=\ww\linewidth,height=\hh\linewidth]{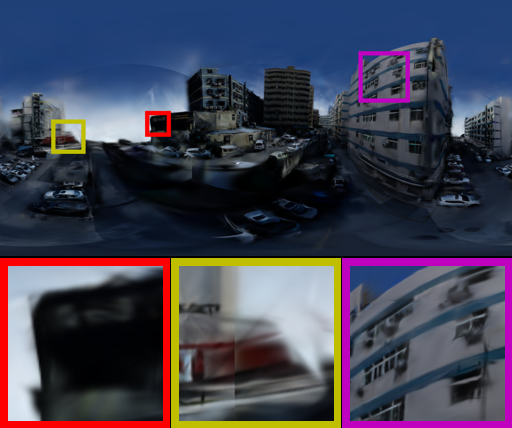}}
    \hfill
    \subfloat[EgoNeRF~\cite{choi2023balanced}]{\includegraphics[width=\ww\linewidth,height=\hh\linewidth]{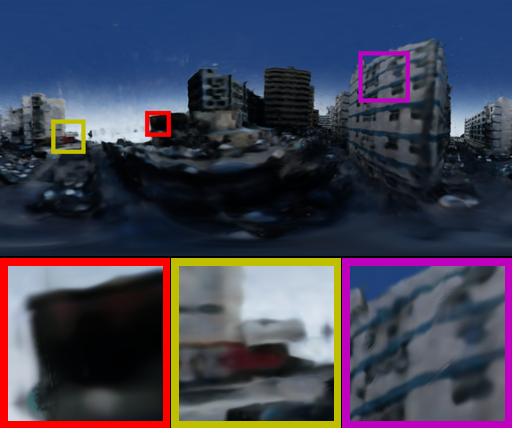}}
    \hfill
    \subfloat[ODGS]{\includegraphics[width=\ww\linewidth,height=\hh\linewidth]{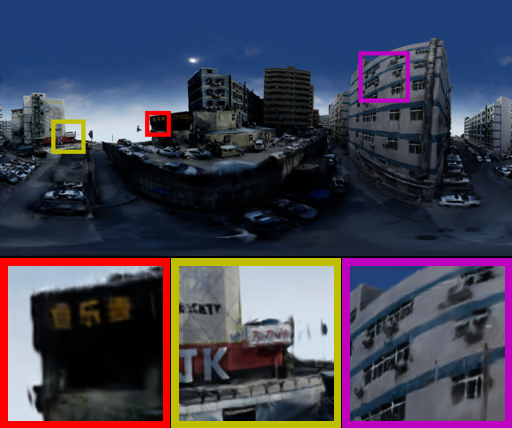}}
    
    \caption{Qualitative comparisons in the roaming scenes (10 min.).
    Each scene is brought from 360Roam, OmniScenes, and 360VO, respectively.
    \textit{Best viewed when zoomed in.}
    }
    \label{fig:qual_roam}
    \vspace{-4mm}
\end{figure}

EgoNeRF shows comparable PSNR with ours in \textit{Ballintoy}, but needs a long optimization time.
We attribute the fast optimization of ODGS to two aspects.
First, while NeRF-based methods use an implicit representation that embeds the scene into a neural network, 3DGS employs explicit representation and directly moves or morphs the elements to optimize the model.
Also, 3DGS exploits the position of SfM point clouds, which can serve as a good initialization point for optimizing Gaussian splats.
3DGS (P), on the other hand, shows high vulnerability to overfitting.
We believe the phenomenon happens because of the weak correlation among the six faces of the cubemap after decomposition.
Since there is no overlap between the six faces, 3DGS is optimized six times independently for faces facing the same direction.
Therefore, even with the same input, the amount of information used is significantly reduced, causing overfitting to occur quickly.

\paragraph{Qualitative comparison}

We also visually compare our method with the other methods in various scenes.
\figref{qual_ego} shows the samples of reconstructed images from egocentric datasets.
We note that the images in the figure are rendered at 10 minutes of training.
The images from EgoNeRF are blurry and contain some artifacts, such as stripe lines or checkerboard patterns, which appear prominent near the edges.
The model is not sufficiently optimized to render the sharp image details.
3DGS trained with the cubemap perspective images sometimes show sharp reconstruction contents, such as a cubic pattern of a frame in the middle row (yellow boundary) but often include unintended projected Gaussian splats that cause image distortion.
We attribute the phenomenon to the rapid overfitting properties of perspective 3DGS.
In contrast, our model successfully reconstructs sharp details in the images.
The superiority of ODGS becomes more noticeable in the roaming dataset, as shown in \figref{qual_roam}.
Although EgoNeRF proposes a balanced grid for egocentric video, it cannot maintain a uniform ray density for every grid if the camera wanders inside a large environment.
As a result, the scene pattern is often completely lost, creating completely different results, and the overall reconstruction quality deteriorates.
While perspective 3DGS shows better quality than EgoNeRF, it often misses some objects or structures where the adjacent faces of the cube meet.
For instance, in the top row, there is an inverted Y-shaped artifact instead of a chair in a purple patch.
This happens because the chair is located where the three sides of the cube meet, and the object is not made from any of the sides.
ODGS overcomes the challenge by optimizing the Gaussian using the whole image and showing prominent performance on both egocentric and roaming datasets.

\paragraph{Ablation Study: Dynamic Densification Strategy for Omnidirectional Images}
\begin{figure}[ht!]
    \centering
    \subfloat[Ground truth(full)]{\includegraphics[width=0.245\linewidth]{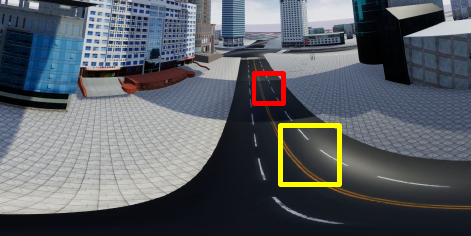}}
    \hfill
    \subfloat[Ground truth]{\includegraphics[width=0.245\linewidth]{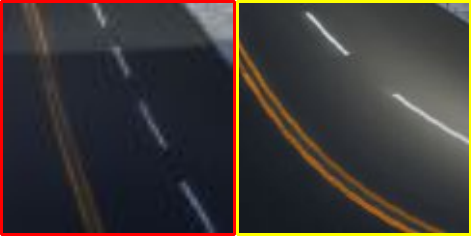}}
    \hfill
    \subfloat[Static threshold]{\includegraphics[width=0.245\linewidth]{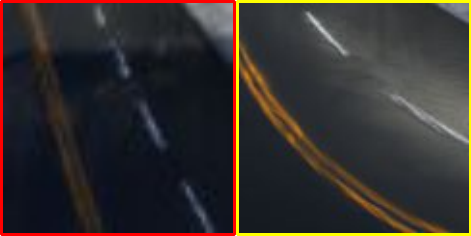}}
    \hfill
    \subfloat[Dynamic threshold]{\includegraphics[width=0.245\linewidth]{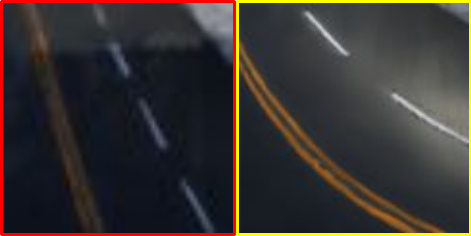}}
    \caption{
        Qualitative comparison of rendered images according to the Gaussian densification policy during optimization.
    }
\label{fig:densification}
\end{figure}
We qualitatively compare and display the results in Figure~\ref{fig:densification} when applying the proposed dynamic densification rule proposed in Section~\ref{sec:densification}.
As shown in the figure, the lanes appear split, with artifact-like patterns emerging on the road due to static densification, as employed in the original 3DGS work~\cite{kerbl20233d}.
Conversely, our densification strategy significantly enhances the model's representation power, leading to markedly more accurate rasterization results.

\section{Conclusion}

In this work, we propose a new method called ODGS, specifically designed to reconstruct 3D scenes from omnidirectional images using 3D Gaussian splatting.
To optimize 3D Gaussian splatting in the omnidirectional image domain, we introduce a new rasterizer that appropriately models the equirectangular projection from the 3D space to the image.
Specifically, we define a tangent plane for each Gaussian and project the Gaussian into the plane, followed by horizontal stretching and rescaling to the pixel space.
Compared to the state-of-the-art NeRF-based methods, ODGS shows about 100 times faster optimization and rendering speed, which allows the user to synthesize the novel view in real-time.
Furthermore, ODGS shows the best reconstruction performance for various input images, including egocentric and roaming scenes, indoors and outdoors.

\paragraph{Limitations and future work}
ODGS still relies on local affine approximation when projecting a Gaussian splat to the camera surface.
Equirectangular projection is not a linear transformation, and straight lines in the 3D space should be expressed as curves in the omnidirectional image.
However, a 3D Gaussian is approximated as a 2D Gaussian, leading to errors that produce artifacts in the rendered image.
Adopting a more accurate distribution for spherically projected Gaussians can reduce errors and enhance the efficiency of the framework.

\clearpage
\section*{Acknowledgements}
This work was supported in part by the IITP grants [No.2021-0-01343, Artificial Intelligence Graduate School Program (Seoul National University), No. 2021-0-02068, and No.2023-0-00156], the NRF grant [No. 2021M3A9E4080782] funded by the Korea government (MSIT), and the SNU-LG AI Research Center.

{
    \small
    \bibliographystyle{ieeenat_fullname}
    \bibliography{ref}
}

\clearpage
\appendix

\section{Appendix / supplemental material}

\subsection{More Implementation Details}
\label{app:parameters}
We follow the hyper-parameters of original 3DGS~\cite{kerbl20233d} excluding some hyperparameters.
Firstly, we set \textit{iterations} as 200k, \textit{densify\_until\_iter} as 100k.
However, we stopped the optimization after 100 minutes, regardless of the current iteration.
For densification we set \textit{percent\_dense} as 1e-3, \textit{densify\_grad\_threshold\_min}~$(\tau_{\text{min}})$ as 2e-5, and \textit{densify\_grad\_threshold\_max}~$(\tau_{\text{max}})$ as 1e-4.

\subsection{Proof of Mathematical Equivalence of the Derived Method}
\label{app:proof}

Here, we present the direct derivation of \eqnref{omni_Jacobian} by differentiating the omnidirectional projection function $\pi_0$ from \eqnref{equirec_proj}.

\begin{align}
\begin{split}
    \frac{\partial \pi_{o}\left(\bm{\mu}\right)}{\partial \bm{\mu}} &= 
    \begin{bmatrix}
    \frac{W}{2\pi}\frac{\bm{\mu}_z}{\bm{\mu}_x^2+\bm{\mu}_z^2} & 0 & -\frac{W}{2\pi}\frac{\bm{\mu}_x}{\bm{\mu}_x^2+\bm{\mu}_z^2} \\
    \\
    -\frac{H}{\pi||\bm{\mu}||^2}\frac{\bm{\mu}_x\bm{\mu}_y}{\sqrt{\bm{\mu}_x^2+\bm{\mu}_z^2}} & \frac{H}{\pi||\bm{\mu}||^2}\sqrt{\bm{\mu}_x^2+\bm{\mu}_z^2} & -\frac{H}{\pi||\bm{\mu}||^2}\frac{\bm{\mu}_y\bm{\mu}_z}{\sqrt{\bm{\mu}_x^2+\bm{\mu}_z^2}} \\
    \end{bmatrix} \\
    \\
    &= \begin{bmatrix}
    \frac{W}{2\pi||\bm{\mu}||} \frac{||\bm{\mu}||}{\sqrt{\bm{\mu}_x^2+\bm{\mu}_z^2}}\frac{\bm{\mu}_z}{\sqrt{\bm{\mu}_x^2+\bm{\mu}_z^2}} & 0 & -\frac{W}{2\pi||\bm{\mu}||}\frac{||\bm{\mu}||}{\sqrt{\bm{\mu}_x^2+\bm{\mu}_z^2}}\frac{\bm{\mu}_x}{\sqrt{\bm{\mu}_x^2+\bm{\mu}_z^2}} \\
    \\
    \frac{H}{\pi||\bm{\mu}||}\frac{-\bm{\mu}_y}{||\bm{\mu}||}\frac{\bm{\mu}_x}{\sqrt{\bm{\mu}_x^2+\bm{\mu}_z^2}} & \frac{H}{\pi||\bm{\mu}||}\frac{\sqrt{\bm{\mu}_x^2+\bm{\mu}_z^2}}{||\bm{\mu}||} & \frac{H}{\pi||\bm{\mu}||}\frac{-\bm{\mu}_y}{||\bm{\mu}||}\frac{\bm{\mu}_z}{\sqrt{\bm{\mu}_x^2+\bm{\mu}_z^2}} \\
    \end{bmatrix} \\
    \\
    &= \begin{bmatrix}
    \frac{W}{2\pi ||\bm{\mu}||}\sec{\theta_\mu}\cos{\phi_\mu} & 0 & -\frac{W}{2\pi ||\bm{\mu}||} \sec{\theta_\mu}\sin{\phi_\mu} \\
    \\
    \frac{H}{\pi||\bm{\mu}||}\sin{\theta_\mu}\sin{\phi_\mu} & \frac{H}{\pi||\bm{\mu}||}\cos{\theta_\mu} & \frac{H}{\pi||\bm{\mu}||}\sin{\theta_\mu}\cos{\phi_\mu} \\
    \end{bmatrix} \\
    \\
    &= \mathbf{J}_{omni}
    \label{eq:jacobian}
\end{split}
\end{align}

This proof demonstrates the mathematical correctness of our description outlined through \eqnref{omni_rotation}, \eqnref{omni_J}, and \eqnref{omni_QS}.
The description in the main paper reveals the underlying assumptions (local affine approximation, tangent plane to sphere surface) and confirms their mathematical validity.

\subsection{Back-Propagation of Rasterization in omnidirectional Image Domain}
\label{app:backprop}

The gradient computation from Gaussian covariance is related to \eqnref{jacobian}.
We denote the gradient value matrix for the projected 2D covariance matrix($\Sigma$) as $\frac{\partial L}{\partial \Sigma}$. 
The size of $\frac{\partial L}{\partial \Sigma}$ is $2\times2$, the same as the original covariance matrix.
Note that the values of $\frac{\partial L}{\partial \Sigma}_{(1,2)}$ and $\frac{\partial L}{\partial \Sigma}_{(2,1)}$ are same since both $\Sigma$ and $\frac{\partial L}{\partial \Sigma}$ are symmetric matrices.

We define $\mathbf{T}$ as $\mathbf{JW}$.
The gradient value matrix of $\mathbf{T}$ is computed as below:

\begin{equation}
\begin{split}
    \frac{\partial \mathcal{L}}{\partial \mathbf{T}}_{(1,1)} &=  2\left(T_{(1,1)} V_{(1,1)} + T_{(1,2)} V_{(1,2)} + T_{(1,3)} V_{(1,3)}\right) * \frac{\partial \mathcal{L}}{\partial \mathbf{\Sigma}}_{(1,1)} \\
    &+ \left(T_{(2,1)} V_{(1,1)} + T_{(2,2)} V_{(1,2)} + T_{(2,3)} V_{(1,3)}\right) * \frac{\partial \mathcal{L}}{\partial \mathbf{\Sigma}}_{(1,2)}, \\
    \frac{\partial \mathcal{L}}{\partial \mathbf{T}}_{(1,2)} &=  2\left(T_{(1,1)} V_{(2,1)} + T_{(1,2)} V_{(2,2)} + T_{(1,3)} V_{(2,3)}\right) * \frac{\partial L}{\partial \mathbf{\Sigma}}_{(1,1)} \\
    &+ \left(T_{(2,1)} V_{(2,1)} + T_{(2,2)} V_{(2,2)} + T_{(2,3)} V_{(2,3)}\right) * \frac{\partial L}{\partial \mathbf{\Sigma}}_{(1,2)}, \\
    \frac{\partial \mathcal{L}}{\partial \mathbf{T}}_{(1,3)} &=  2\left(T_{(1,1)} V_{(3,1)} + T_{(1,2)} V_{(3,2)} + T_{(1,3)} V_{(3,3)}\right) * \frac{\partial L}{\partial \mathbf{\Sigma}}_{(1,1)} \\
    &+ \left(T_{(2,1)} V_{(3,1)} + T_{(2,2)} V_{(3,2)} + T_{(2,3)} V_{(3,3)}\right) * \frac{\partial L}{\partial \mathbf{\Sigma}}_{(1,2)}, \\
    \frac{\partial \mathcal{L}}{\partial \mathbf{T}}_{(2,1)} &=  2\left(T_{(2,1)} V_{(1,1)} + T_{(2,2)} V_{(1,2)} + T_{(2,3)} V_{(1,3)}\right) * \frac{\partial \mathcal{L}}{\partial \mathbf{\Sigma}}_{(2,2)} \\
    &+ \left(T_{(1,1)} V_{(1,1)} + T_{(1,2)} V_{(1,2)} + T_{(1,3)} V_{(1,3)}\right) * \frac{\partial \mathcal{L}}{\partial \mathbf{\Sigma}}_{(1,2)}, \\
    \frac{\partial \mathcal{L}}{\partial \mathbf{T}}_{(2,2)} &=  2\left(T_{(2,1)} V_{(2,1)} + T_{(2,2)} V_{(2,2)} + T_{(2,3)} V_{(2,3)}\right) * \frac{\partial \mathcal{L}}{\partial \mathbf{\Sigma}}_{(2,2)} \\
    &+ \left(T_{(1,1)} V_{(2,1)} + T_{(1,2)} V_{(2,2)} + T_{(1,3)} V_{(2,3)}\right) * \frac{\partial \mathcal{L}}{\partial \mathbf{\Sigma}}_{(1,2)}, \\
    \frac{\partial \mathcal{L}}{\partial \mathbf{T}}_{(2,3)} &=  2\left(T_{(2,1)} V_{(3,1)} + T_{(2,2)} V_{(3,2)} + T_{(2,3)} V_{(3,3)}\right) * \frac{\partial \mathcal{L}}{\partial \mathbf{\Sigma}}_{(2,2)} \\
    &+ \left(T_{(1,1)} V_{(3,1)} + T_{(1,2)} V_{(3,2)} + T_{(1,3)} V_{(3,3)}\right) * \frac{\partial \mathcal{L}}{\partial \mathbf{\Sigma}}_{(1,2)}.
\end{split}
\label{eq:backprop_T}
\end{equation}

Then, the gradient for Jacobian matrix, $\frac{\partial \mathcal{L}}{\partial \mathbf{J}}$, is calculated as multiplication of $\mathbf{W}^T$ and $\frac{\partial \mathcal{L}}{\partial \mathbf{T}}$.
After the gradient of $\mathbf{J}$ is calculated, the gradient for each position is computed as follows:

\begin{equation}
    \begin{split}
    \frac{\partial \mathcal{L}}{\partial t_x} = &- \frac{W}{\pi} \cdot \frac{t_x t_z}{\left(t^2_x+t^2_z\right)^2} \cdot \frac{\partial \mathcal{L}}{\partial \mathbf{J}}_{(1,1)} + \frac{W}{2\pi} \cdot \frac{t^2_x-t^2_z}{\left(t^2_x+t^2_z\right)^2} \cdot \frac{\partial \mathcal{L}}{\partial \mathbf{J}}_{(1,3)} \\
    &+ \frac{H}{\pi} \cdot \frac{t_y \left( t^2_z t^2_r - 2t^2_x\left(t^2_x+t^2_z\right) \right)}{t^4_r \left(t^2_x+t^2_z\right)^{3/2}} \cdot \frac{\partial \mathcal{L}}{\partial \mathbf{J}}_{(2,1)} + \frac{H}{\pi} \cdot \frac{t_x \left( t^2_r - 2t^2_y \right)}{t^4_r \sqrt{t^2_x+t^2_z}} \cdot \frac{\partial \mathcal{L}}{\partial \mathbf{J}}_{(2,2)} \\
    &- \frac{H}{\pi} \cdot \frac{t_x t_y t_z \left( 2\left(t^2_x+t^2_z\right) + t^2_r \right)}{t^4_r \left(t^2_x+t^2_z\right)^{3/2}} \cdot \frac{\partial \mathcal{L}}{\partial \mathbf{J}}_{(2,3)}. \\
    \frac{\partial \mathcal{L}}{\partial t_y} = &+\frac{H}{\pi} \cdot \frac{t_x \left( t^2_r - 2t^2_y \right)}{t^4_r \sqrt{t^2_x+t^2_z}} \cdot \frac{\partial \mathcal{L}}{\partial \mathbf{J}}_{(2,1)} + \frac{2H}{\pi} \cdot \frac{t_y \sqrt{t^2_x+t^2_z}}{t^4_r} \cdot \frac{\partial \mathcal{L}}{\partial \mathbf{J}}_{(2,2)} \\
    &+ \frac{H}{\pi} \cdot \frac{t_z \left( t^2_r - 2t^2_y \right)}{t^4_r \sqrt{t^2_x+t^2_z}} \cdot \frac{\partial \mathcal{L}}{\partial \mathbf{J}}_{(2,3)}. \\
    \frac{\partial \mathcal{L}}{\partial t_z} = &+\frac{W}{2\pi} \cdot \frac{t^2_x-t^2_z}{\left(t^2_x+t^2_z\right)^2} \cdot \frac{\partial \mathcal{L}}{\partial \mathbf{J}}_{(1,1)} + \frac{W}{\pi} \cdot \frac{t_x t_z}{\left(t^2_x+t^2_z\right)^2} \cdot \frac{\partial \mathcal{L}}{\partial \mathbf{J}}_{(1,3)} \\
    &- \frac{H}{\pi} \cdot \frac{t_x t_y t_z \left( 2\left(t^2_x+t^2_z\right) + t^2_r \right)}{t^4_r \left(t^2_x+t^2_z\right)^{3/2}} \cdot \frac{\partial \mathcal{L}}{\partial \mathbf{J}}_{(2,1)} + \frac{H}{\pi} \cdot \frac{t_z \left( t^2_r - 2t^2_y \right)}{t^4_r \sqrt{t^2_x+t^2_z}} \cdot \frac{\partial \mathcal{L}}{\partial \mathbf{J}}_{(2,2)} \\
    &+ \frac{H}{\pi} \cdot \frac{t_y \left( t^2_x t^2_r - 2t^2_z\left(t^2_x+t^2_z\right) \right)}{t^4_r \left(t^2_x+t^2_z\right)^{3/2}} \cdot \frac{\partial \mathcal{L}}{\partial \mathbf{J}}_{(2,3)}.
    \end{split}
\end{equation}

\newpage

\subsection{Comparison with Original 3DGS with more input images.}
\begin{table}[h]
    \centering
    \captionof{table}{Quantitative comparison of 3DGS (P) in 6-views and 18-views.}
    \begin{tabular}{c|c|ccc|ccc}
        \toprule[1.0pt]
        \multirow{2}{*}{Dataset} & \multirow{2}{*}{Method} & \multicolumn{3}{c|}{10 min} & \multicolumn{3}{c}{100 min} \\
        & & PSNR$_{\uparrow}$ & SSIM$_{\uparrow}$ & LPIPS$_{\downarrow}$ & PSNR$_{\uparrow}$ & SSIM$_{\uparrow}$ & LPIPS$_{\downarrow}$ \\
        \midrule
        \multirow{3}{*}{OmniBlender}
        & 3DGS (P6)  & 29.36 & 0.8770 & 0.1400 & 21.19 & 0.7528 & 0.3021 \\
        & 3DGS (P18) & 27.85 & 0.8387 & 0.1737 & 24.56 & 0.7907 & 0.2478 \\
        & ODGS & \textbf{32.76} & \textbf{0.9234} & \textbf{0.0469} & \textbf{33.05} & \textbf{0.9229} & \textbf{0.0343} \\
        \midrule
        \multirow{3}{*}{Ricoh360}
        & 3DGS (P6)  & \textbf{25.12} & 0.7932 & 0.2397 & 22.07 & 0.7228 & 0.3218 \\
        & 3DGS (P18) & 24.76 & 0.7726 & 0.2565 & 23.14 & 0.7277 & 0.3109 \\
        & ODGS & 24.94 & \textbf{0.8135} & \textbf{0.1489} & \textbf{26.27} & \textbf{0.8462} & \textbf{0.1051} \\
        \midrule
        \multirow{3}{*}{OmniPhotos}
        & 3DGS (P6)  & 25.61 & 0.8310 & 0.2100 & 23.30 & 0.7859 & 0.2670 \\
        & 3DGS (P18) & 24.93 & 0.8007 & 0.2412 & 23.21 & 0.7541 & 0.2996 \\
        & ODGS & \textbf{26.24} & \textbf{0.8704} & \textbf{0.1108} & \textbf{27.04} & \textbf{0.8878} & \textbf{0.0875} \\
        \midrule
        \midrule
        \multirow{3}{*}{360Roam}
        & 3DGS (P6)  & 20.17 & 0.7001 & 0.3536 & 19.34 & 0.6576 & 0.3837 \\
        & 3DGS (P18) & 20.88 & 0.6992 & 0.3571 & \textbf{21.05} & 0.6994 & 0.3405 \\
        & ODGS & \textbf{21.08} & \textbf{0.7066} & \textbf{0.3003} & 20.85 & \textbf{0.7111} & \textbf{0.2254} \\
        \midrule
        \multirow{3}{*}{OmniScenes}
        & 3DGS (P6)  & 23.61 & 0.8444 & 0.2835 & 17.14 & 0.7119 & 0.3906 \\
        & 3DGS (P18) & 24.00 & 0.8400 & 0.1993 & 21.43 & 0.7864 & 0.2828 \\
        & ODGS & \textbf{24.42} & \textbf{0.8526} & \textbf{0.1391} & \textbf{24.51} & \textbf{0.8505} & \textbf{0.1282} \\
        \midrule
        \multirow{3}{*}{360VO}
        & 3DGS (P6)  & 22.87 & 0.7861 & 0.2970 & 22.73 & 0.7822 & 0.3061 \\
        & 3DGS (P18) & 23.22 & 0.7875 & 0.2939 & 23.57 & 0.7938 & 0.2825 \\
        & ODGS & \textbf{24.63} & \textbf{0.8245} & \textbf{0.2175} & \textbf{26.68} & \textbf{0.8694} & \textbf{0.1264} \\
        \bottomrule
    \end{tabular}
    \label{tab:more_persp}
\end{table}

Although we report the results of the original 3DGS method using 6 cubemap decomposed perspective images in \tabref{comp1}, we also compare the results when 3DGS is optimized with more perspective images.
Whereas the original decomposition generates six perspective images for one omnidirectional image, the new decomposition produces 18 images by adding 12 perspective cameras where each camera faces an edge of the cube.
\tabref{more_persp} shows the performance of optimized results according to the number of perspective images for optimization.
When using the 18 views (P18), the performance is comparable to the 6 views (P6) at the 10-minute mark but surpasses the 6 view results after the 100-minute optimization.
At first, the increased number of images for training prevents the model from sufficiently learning from all views in the early stages (10 minutes), resulting in slightly lower performance.
However, after sufficient optimization time (100 minutes) passes, the additional views allow for further optimization, leading to improved results.
Still, ODGS shows the highest performance in most metrics, even considering 3DGS using 18 views, demonstrating the superiority of our rasterizer.

\newpage

\subsection{Detailed Quantitative Results}

We provide detailed quantitative results that compose \tabref{comp1} below.
We report PSNR, SSIM, and LPIPS of each method for all scenes in \tabref{omniblender_performance_comparison} - \tabref{360vo_performance_comparison}.
The tables show that our method outperforms all the baselines in almost all scenes.
Please note that the \textit{room1} scene in the OmniScenes dataset was omitted because the OpenMVG did not function properly due to the image sequence jumping in the middle.
In \tabref{comp1}, we report the averages of the scenes, excluding room1.

\begin{table}[h]
    \centering
    \caption{Quantitative comparison of 3D reconstruction results on Omniblender dataset (10min/100min).
    The best result for each metric is written in \textbf{bold}.
    Our method shows the best performance on almost all settings.}
    \resizebox{\textwidth}{!}{
    \begin{tabular}{c|c|c|c|c|c|c}
        \toprule[1.0pt]
        opt time & scene & Nerf(P) & 3DGS(P) & TensoRF & EgoNeRF & ODGS \\
        \midrule
        \multirow{12}{*}{10min} & archiviz-flat & 20.42 / 0.7431 / 0.4028 & 31.20 / 0.9158 / 0.1044 & 28.44 / 0.8406 / 0.2751 & 29.06 / 0.8501 / 0.2477 & \textbf{34.47} / \textbf{0.9499} / \textbf{0.0246} \\
         & barbershop & 19.50 / 0.6816 / 0.5540 & 33.25 / 0.9472 / 0.0849 & 27.58 / 0.8289 / 0.3259 & 31.13 / 0.9065 / 0.1793 & \textbf{38.16} / \textbf{0.9778} / \textbf{0.0203} \\
         & bistro bike & 17.49 / 0.5145 / 0.6151 & 30.56 / 0.9298 / 0.0890 & 21.81 / 0.5978 / 0.4968 & 30.01 / 0.9065 / 0.1035 & \textbf{33.87} / \textbf{0.9712} / \textbf{0.0195} \\
         & bistro square & 16.27 / 0.4960 / 0.6003 & 24.95 / 0.8887 / 0.1273 & 19.52 / 0.5620 / 0.4974 & 23.84 / 0.8298 / 0.1605 & \textbf{28.59} / \textbf{0.9500} / \textbf{0.0269} \\
         & classroom & 18.16 / 0.6125 / 0.6144 & 26.89 / 0.8109 / 0.2291 & 25.02 / 0.7206 / 0.5100 & 26.33 / 0.7662 / 0.3740 & \textbf{31.54} / \textbf{0.8674} / \textbf{0.1161} \\
         & fisher hut & 23.43 / 0.7171 / 0.4084 & 29.69 / 0.8153 / 0.1809 & 28.74 / 0.7549 / 0.4224 & 29.73 / 0.7780 / 0.3098 & \textbf{32.39} / \textbf{0.8551} / \textbf{0.0610} \\
         & lone monk & 16.88 / 0.5561 / 0.5039 & 30.01 / 0.9204 / 0.1029 & 23.18 / 0.6933 / 0.3530 & 28.30 / 0.8777 / 0.1433 & \textbf{33.49} / \textbf{0.9638} / \textbf{0.0201} \\
         & LOU & 17.79 / 0.6211 / 0.4705 & 30.68 / 0.9309 / 0.1030 & 27.60 / 0.8651 / 0.2174 & 30.89 / 0.9083 / 0.1171 & \textbf{35.17} / \textbf{0.9573} / \textbf{0.0314} \\
         & pavilion midday chair & 19.99 / 0.7027 / 0.5128 & 29.69 / 0.9127 / 0.1015 & 26.33 / 0.7850 / 0.3237 & 29.01 / 0.8827 / 0.1354 & \textbf{32.13} / \textbf{0.9499} / \textbf{0.0324} \\
         & pavilion midday pond & 17.86 / 0.5496 / 0.5690 & 24.82 / 0.7914 / 0.1797 & 22.11 / 0.6613 / 0.3091 & 23.86 / 0.7411 / 0.1963 & \textbf{25.23} / \textbf{0.8125} / \textbf{0.0954} \\
         & restroom & 23.37 / 0.5421 / 0.6432 & 31.20 / 0.7840 / 0.2368 & 28.60 / 0.6644 / 0.5093 & 29.07 / 0.6930 / 0.4460 & \textbf{35.37} / \textbf{0.9020} / \textbf{0.0684} \\
         \cmidrule{2-7}
         & average & 19.20 / 0.6124 / 0.5359 & 29.36 / 0.8770 / 0.1400 & 25.36 / 0.7249 / 0.3855 & 28.29 / 0.8309 / 0.2194 & \textbf{32.76} / \textbf{0.9234} / \textbf{0.0469} \\
        \midrule
        \multirow{12}{*}{100min} & archiviz-flat & 21.28 / 0.7326 / 0.3993 & 21.61 / 0.8054 / 0.2724 & 29.02 / 0.8504 / 0.2185 & 32.63 / 0.9175 / 0.1077 & \textbf{34.10} / \textbf{0.9454} / \textbf{0.0243} \\
         & barbershop & 19.82 / 0.6584 / 0.5226 & 22.31 / 0.7708 / 0.3194 & 28.27 / 0.8479 / 0.2606 & 34.48 / 0.9551 / 0.0835 & \textbf{37.89} / \textbf{0.9762} / \textbf{0.0204} \\
         & bistro bike & 18.12 / 0.5077 / 0.5828 & 19.39 / 0.7899 / 0.2414 & 22.58 / 0.6223 / 0.4310 & 33.23 / 0.9546 / 0.0471 & \textbf{36.16} / \textbf{0.9752} / \textbf{0.0144} \\
         & bistro square & 17.06 / 0.5016 / 0.5616 & 21.27 / 0.8010 / 0.2414 & 19.86 / 0.5714 / 0.4319 & 25.49 / 0.8990 / 0.0940 & \textbf{29.32} / \textbf{0.9536} / \textbf{0.0228} \\
         & classroom & 18.87 / 0.5968 / 0.5367 & 18.52 / 0.6499 / 0.4318 & 26.95 / 0.7481 / 0.4384 & 29.55 / 0.8306 / 0.2652 & \textbf{31.54} / \textbf{0.8524} / \textbf{0.0567} \\
         & fisher hut & 24.87 / 0.7221 / 0.3964 & 25.92 / 0.7872 / 0.2380 & 28.74 / 0.7514 / 0.3566 & 30.32 / 0.8018 / 0.2338 & \textbf{32.75} / \textbf{0.8604} / \textbf{0.0525} \\
         & lone monk & 17.27 / 0.5459 / 0.4828 & 19.12 / 0.7479 / 0.3114 & 23.73 / 0.7161 / 0.2889 & 30.90 / 0.9311 / 0.0800 & \textbf{32.88} / \textbf{0.9608} / \textbf{0.0180} \\
         & LOU & 19.93 / 0.6619 / 0.4103 & 21.59 / 0.8199 / 0.2515 & 28.60 / 0.8430 / 0.1756 & 33.53 / 0.9400 / 0.0693 & \textbf{35.43} / \textbf{0.9573} / \textbf{0.0277} \\
         & pavilion midday chair & 21.38 / 0.6994 / 0.4329 & 19.62 / 0.7522 / 0.3360 & 27.11 / 0.8015 / 0.2558 & 31.08 / 0.9315 / 0.0648 & \textbf{32.93} / \textbf{0.9562} / \textbf{0.0266} \\
         & pavilion midday pond & 17.60 / 0.5346 / 0.5142 & 20.89 / 0.7267 / 0.2596 & 22.59 / 0.6795 / 0.2512 & 25.71 / 0.8145 / 0.1240 & \textbf{25.40} / \textbf{0.8189} / \textbf{0.0783} \\
         & restroom & 24.20 / 0.5407 / 0.6039 & 22.91 / 0.6301 / 0.4204 & 29.49 / 0.7261 / 0.3787 & 32.86 / 0.8515 / 0.2167 & \textbf{35.14} / \textbf{0.8957} / \textbf{0.0354} \\
         \cmidrule{2-7}
         & average & 20.04 / 0.6092 / 0.4949 & 21.19 / 0.7528 / 0.3021 & 26.08 / 0.7416 / 0.3170 & 30.89 / 0.8934 / 0.1260 & \textbf{33.05} / \textbf{0.9229} / \textbf{0.0343} \\
        \bottomrule
    \end{tabular}
    }
    \label{tab:omniblender_performance_comparison}
\end{table}
\begin{table}[h]
    \centering
    \caption{Quantitative comparison of 3D reconstruction results on Ricoh360 dataset (10min/100min).
    The best result for each metric is written in \textbf{bold}.
    Our method shows the best performance on almost all settings.
    }
    \resizebox{\textwidth}{!}{
    \begin{tabular}{c|c|c|c|c|c|c}
        \toprule[1.0pt]
        opt. time & scene & NeRF(P) & 3DGS(P) & TensoRF & EgoNeRF & ODGS \\
        \midrule
         \multirow{13}{*}{10min} & bricks & 12.40 / 0.4489 / 0.6403 & 23.88 / 0.7875 / 0.2275 & 21.08 / 0.6201 / 0.4918 & 23.09 / 0.7223 / 0.2984 &  \textbf{23.90}/ \textbf{0.8185} / \textbf{0.1296}\\
         & bridge & 14.96 / 0.5553 / 0.5866 & 23.78 / 0.7783 / 0.2145 & 21.93 / 0.6456 / 0.4823 & 23.34 / 0.7199 / 0.3096 &  \textbf{23.88} / \textbf{0.7987} / \textbf{0.1286}  \\
         & bridge under & 19.71 / 0.4987 / 0.6769 & 24.30 / 0.7892 / 0.2239 & 21.99 / 0.6323 / 0.5791 & 24.11 / 0.7504 / 0.3202 &  \textbf{25.12} / \textbf{0.8347} / \textbf{0.1339} \\
         & cat tower & 12.54 / 0.5182 / 0.5990 & 24.33 / 0.7543 / 0.2548 & 22.45 / 0.6308 / 0.6002 & 23.80 / 0.6861 / 0.3758 & \textbf{24.47} / \textbf{0.7771} / \textbf{0.1435} \\
         & center & 14.76 / 0.6691 / 0.5211 & 27.24 / 0.8364 / 0.2887 & 27.23 / 0.8088 / 0.4294 & 27.97 / 0.8450 / 0.2521 & \textbf{28.10} / \textbf{0.8710} / \textbf{0.1206} \\
         & farm & 14.45 / 0.4970 / 0.6262 & 21.66 / \textbf{0.6897} / 0.3248 & 20.80 / 0.5683 / 0.5141 & \textbf{21.80} / 0.6483 / 0.3386 & 20.74 / 0.6881 / \textbf{0.2270} \\
         & flower & 12.03 / 0.4132 / 0.6912 & 21.71 / 0.6942 / 0.3247 & 20.07 / 0.5414 / 0.6696 & 21.51 / 0.6149 / 0.4211 & \textbf{22.19} / \textbf{0.7273} / \textbf{0.1925} \\
         & gallery chair & 15.40 / 0.6950 / 0.4929 & \textbf{27.76} / 0.8732 / 0.1962 & 26.00 / 0.7907 / 0.5233 & 27.01 / 0.8326 / 0.3409 & 27.29 / \textbf{0.8777} / \textbf{0.1353} \\
         & gallery park & 12.29 / 0.6050 / 0.5176 & 25.30 / 0.8021 / 0.2384 & 24.21 / 0.7394 / 0.5120 & 25.11 / 0.7703 / 0.3245 & \textbf{25.48} / \textbf{0.8241} / \textbf{0.1341} \\
         & gallery pillar & 14.50 / 0.6445 / 0.4902 & 27.79 / 0.8613 / 0.1617 & 25.85 / 0.7821 / 0.3977 & 27.31 / 0.8312 / 0.2379 & \textbf{28.02} / \textbf{0.8821} / \textbf{0.0882} \\
         & garden & 13.97 / 0.5682 / 0.5430 & \textbf{27.53} / \textbf{0.7919} / \textbf{0.2118} & 25.37 / 0.6649 / 0.5616 & 26.48 / 0.7175 / 0.3517 & 23.20 / 0.7843 / 0.2289 \\
         & poster & 14.99 / 0.6258 / 0.5679 & 26.14 / 0.8599 / 0.2098 & 23.20 / 0.7500 / 0.4784 & 25.39 / 0.8213 / 0.3205 & \textbf{26.90} / \textbf{0.8782} / \textbf{0.1249} \\
         \cmidrule{2-7}
         & average & 14.33 / 0.5616 / 0.5794 & \textbf{25.12} / 0.7932 / 0.2397 & 23.35 / 0.6812 / 0.5200 & 24.74 / 0.7467 / 0.3243 & 24.94 / \textbf{0.8135} / \textbf{0.1489} \\
        \midrule
        \multirow{13}{*}{100min} & bricks & 15.01 / 0.4760 / 0.6245 & 22.60 / 0.7410 / 0.2855 & 21.66 / 0.6353 / 0.4375 & 23.93 / 0.7616 / 0.2475 & \textbf{24.62} / \textbf{0.8479} / \textbf{0.1021} \\
         & bridge & 17.32 / 0.5558 / 0.5620 & 21.94 / 0.7157 / 0.3133 & 22.58 / 0.6558 / 0.4306 & 23.94 / 0.7516 / 0.2562 & \textbf{24.37} / \textbf{0.8154} / \textbf{0.1063} \\
         & bridge under & 16.42 / 0.5075 / 0.6447 & 19.03 / 0.6377 / 0.3663 & 22.86 / 0.6577 / 0.4826 & 25.05 / 0.7924 / 0.2492 & \textbf{25.93} / \textbf{0.8538} / \textbf{0.1026} \\
         & cat tower & 15.45 / 0.5323 / 0.5824 & 21.24 / 0.6851 / 0.3565 & 23.02 / 0.6393 / 0.5477 & 24.52 / 0.7163 / 0.3417 & \textbf{25.35} / \textbf{0.8088} / \textbf{0.1109} \\
         & center & 17.09 / 0.6566 / 0.4955 & 20.04 / 0.6974 / 0.4237 & 27.90 / 0.8182 / 0.3840 & 29.12 / 0.8625 / 0.2119 & \textbf{29.39} / \textbf{0.8940} / \textbf{0.0808} \\
         & farm & 15.93 / 0.4830 / 0.6173 & 21.49 / \textbf{0.6844} / 0.3299 & 21.09 / 0.5765 / 0.4662 & \textbf{22.25} / 0.6745 / \textbf{0.3089} & 16.34 / 0.6039 / 0.4109 \\
         & flower & 13.57 / 0.4153 / 0.6845 & 20.48 / 0.6559 / 0.3531 & 20.57 / 0.5506 / 0.6161 & 22.08 / 0.6497 / 0.3922 & \textbf{22.71} / \textbf{0.7485} / \textbf{0.1509} \\
         & gallery chair & 17.59 / 0.6873 / 0.5240 & 26.44 / 0.8509 / 0.2161 & 26.61 / 0.7990 / 0.4766 & \textbf{27.71} / 0.8505 / 0.2993 & 27.62 / \textbf{0.8831} / \textbf{0.1135} \\
         & gallery park & 14.24 / 0.5847 / 0.5404 & 23.22 / 0.7637 / 0.3027 & 24.64 / 0.7457 / 0.4724 & 25.64 / 0.7848 / 0.3001 & \textbf{26.19} / \textbf{0.8401} / \textbf{0.1076} \\
         & gallery pillar & 16.57 / 0.6405 / 0.4946 & 21.93 / 0.7429 / 0.3205 & 26.49 / 0.7960 / 0.3336 & 27.97 / 0.8467 / 0.2104 & \textbf{28.74} / \textbf{0.8970} / \textbf{0.0693} \\
         & garden & 17.81 / 0.5840 / 0.5101 & 25.97 / 0.7792 / 0.2528 & 25.91 / 0.6738 / 0.5201 & \textbf{27.16} / 0.7441 / 0.3112 & 27.09 / \textbf{0.8383} / \textbf{0.1006} \\
         & poster & 16.91 / 0.6169 / 0.5794 & 20.45 / 0.7199 / 0.3406 & 24.32 / 0.7750 / 0.4161 & 26.50 / 0.8497 / 0.2613 & \textbf{26.92} / \textbf{0.8808} / \textbf{0.1113} \\
         \cmidrule{2-7}
         & average & 16.16 / 0.5617 / 0.5716 & 22.07 / 0.7228 / 0.3218 & 23.97 / 0.6936 / 0.4653 & \textbf{25.49} / 0.7737 / 0.2825 & 25.44 / \textbf{0.8260} / \textbf{0.1306} \\
        \bottomrule
    \end{tabular}
    }
    \label{tab:ricoh360_performance_comparison}
\end{table}

\begin{table}[t]
    \centering
    \caption{Quantitative comparison of 3D reconstruction results on Omniphotos dataset (10min/100min).
    The best result for each metric is written in \textbf{bold}.
    Our method shows the best performance on almost all settings.
    }
    \resizebox{\textwidth}{!}{
    \begin{tabular}{c|c|c|c|c|c|c}
        \toprule[1.0pt]
        opt time & scene & Nerf(P) & 3DGS(P) & TensoRF & EgoNeRF & ODGS \\
        \midrule
        \multirow{11}{*}{10min} & Ballintoy & 19.90 / 0.7292 / 0.4717 & 28.67 / 0.8875 / 0.2094 & 25.68 / 0.8008 / 0.4190 & 28.49 / 0.8715 / 0.2270 & \textbf{29.11} / \textbf{0.9085} / \textbf{0.1076} \\
         & BeihaiPark & 16.76 / 0.5946 / 0.5871 & 23.39 / 0.8126 / 0.2600 & 22.16 / 0.6855 / 0.5516 & 24.35 / 0.7755 / 0.2682 & \textbf{25.34} / \textbf{0.8600} / \textbf{0.1409} \\
         & Cathedral & 15.38 / 0.4736 / 0.6851 & 23.01 / 0.7885 / 0.2569 & 19.35 / 0.5638 / 0.5511 & 23.11 / 0.7267 / 0.2898 & \textbf{23.74} / \textbf{0.8394} / \textbf{0.1416} \\
         & Coast & 20.38 / 0.6452 / 0.5253 & 27.86 / 0.8378 / 0.2011 & 24.69 / 0.7026 / 0.4440 & 27.74 / 0.8006 / 0.2320 & \textbf{28.75} / \textbf{0.8837} / \textbf{0.0966} \\
         & Field & 23.44 / 0.7063 / 0.4293 & 29.51 / 0.8392 / 0.1600 & 27.25 / 0.7427 / 0.4602 & 28.53 / 0.7843 / 0.2618 & \textbf{29.71} / \textbf{0.8702} / \textbf{0.0892} \\
         & Nunobiki2 & 18.79 / 0.6182 / 0.5295 & \textbf{24.84} / 0.8017 / 0.2224 & 22.93 / 0.6719 / 0.5308 & 23.45 / 0.7052 / 0.3378 & 21.62 / \textbf{0.8289} / \textbf{0.1496} \\
         & SecretGarden1 & 17.96 / 0.6380 / 0.5040 & 25.48 / 0.8579 / 0.1706 & 22.49 / 0.7065 / 0.5205 & 24.87 / 0.7885 / 0.2450 & \textbf{27.53} / \textbf{0.8940} / \textbf{0.0769} \\
         & Shrines1 & 15.67 / 0.4516 / 0.7258 & 22.36 / 0.7449 / 0.2864 & 19.52 / 0.5179 / 0.6441 & 21.28 / 0.6374 / 0.3693 & \textbf{23.45} / \textbf{0.8108} / \textbf{0.1560} \\
         & Temple3 & 15.27 / 0.5932 / 0.6112 & 25.17 / 0.8549 / 0.1839 & 20.83 / 0.6800 / 0.5691 & 24.31 / 0.7916 / 0.2321 & \textbf{26.14} / \textbf{0.8881} / \textbf{0.0868} \\
         & Wulongting & 17.89 / 0.7083 / 0.4451 & 25.80 / 0.8845 / 0.1497 & 22.89 / 0.7697 / 0.3987 & 25.89 / 0.8405 / 0.1991 & \textbf{26.98} / \textbf{0.9203} / \textbf{0.0626} \\
         \cmidrule{2-7}
         & average & 18.14 / 0.6158 / 0.5514 & 25.61 / 0.8310 / 0.2100 & 22.78 / 0.6841 / 0.5089 & 25.20 / 0.7722 / 0.2662 & \textbf{26.24} / \textbf{0.8704} / \textbf{0.1108} \\
        \midrule
         \multirow{11}{*}{100min} & Ballintoy & 23.32 / 0.7538 / 0.3796 & 26.37 / 0.8694 / 0.2286 & 26.85 / 0.8193 / 0.3738 & 29.83 / 0.8981 / 0.1827 & \textbf{30.06} / \textbf{0.9220} / \textbf{0.0780} \\
         & BeihaiPark & 18.48 / 0.6089 / 0.4949 & 20.81 / 0.7581 / 0.2898 & 23.13 / 0.7095 / 0.4783 & \textbf{26.19} / \textbf{0.8390} / \textbf{0.1778} & 19.94 / 0.7955 / 0.2352 \\
         & Cathedral & 18.01 / 0.5032 / 0.6033 & 20.18 / 0.6879 / 0.3682 & 20.23 / 0.5935 / 0.4848 & 24.88 / 0.8056 / 0.1878 & \textbf{25.39} / \textbf{0.8769} / \textbf{0.1081} \\
         & Coast & 23.12 / 0.6666 / 0.4263 & 27.29 / 0.8334 / 0.2091 & 26.27 / 0.7329 / 0.3868 & \textbf{29.28} / 0.8553 / 0.1627 & 29.23 / \textbf{0.9004} / \textbf{0.0770} \\
         & Field & 25.81 / 0.7274 / 0.3980 & 29.36 / 0.8420 / 0.1665 & 27.72 / 0.7488 / 0.4209 & \textbf{30.04} / 0.8374 / 0.1776 & 29.18 / \textbf{0.8866} / \textbf{0.0851} \\
         & Nunobiki2 & 21.02 / 0.6346 / 0.4705 & 20.30 / 0.7116 / 0.3662 & 23.63 / 0.6866 / 0.4686 & 25.13 / 0.7879 / 0.2115 & \textbf{25.78} / \textbf{0.8659} / \textbf{0.0996} \\
         & SecretGarden1 & 20.46 / 0.6573 / 0.4641 & 23.93 / 0.8470 / 0.2015 & 23.39 / 0.7273 / 0.4548 & 26.76 / 0.8514 / 0.1513 & \textbf{27.54} / \textbf{0.8963} / \textbf{0.0725} \\
         & Shrines1 & 18.13 / 0.4803 / 0.6417 & 21.32 / 0.7266 / 0.2910 & 20.01 / 0.5285 / 0.5717 & 22.83 / 0.7313 / 0.2495 & \textbf{23.94} / \textbf{0.8270} / \textbf{0.1319} \\
         & Temple3 & 18.69 / 0.6232 / 0.5202 & 22.22 / 0.7889 / 0.2523 & 22.06 / 0.7041 / 0.4840 & \textbf{26.21} / 0.8518 / 0.1411 & 24.58 / \textbf{0.8895} / \textbf{0.0822} \\
         & Wulongting & 20.98 / 0.7328 / 0.3736 & 21.23 / 0.7945 / 0.2964 & 24.02 / 0.7875 / 0.3434 & \textbf{27.84} / 0.8914 / 0.1240 & 27.66 / \textbf{0.9255} / \textbf{0.0531} \\
         \cmidrule{2-7}
         & average & 20.80 / 0.6388 / 0.4772 & 23.30 / 0.7859 / 0.2670 & 23.73 / 0.7038 / 0.4467 & \textbf{26.90} / 0.8349 / 0.1766 & 26.33 / \textbf{0.8786} / \textbf{0.1023} \\
        \bottomrule
    \end{tabular}
    }
    \label{tab:omniphotos_performance_comparison}
\end{table}
\begin{table}[t]
    \centering
    \caption{Quantitative comparison of 3D reconstruction results on 360Roam dataset (10min/100min).
    The best result for each metric is written in \textbf{bold}.
    Our method shows the best performance on almost all settings.
    }
    \resizebox{\textwidth}{!}{
    \begin{tabular}{c|c|c|c|c|c|c}
        \toprule[1.0pt]
        opt time & scene & Nerf(P) & 3DGS(P) & TensoRF & EgoNeRF & ODGS \\
        \midrule
        \multirow{12}{*}{10min} & bar & 13.49 / 0.6218 / 0.5246 & 18.75 / \textbf{0.6982} / 0.3347 & 15.98 / 0.5212 / 0.7717 & 18.34 / 0.5713 / 0.4191 & \textbf{19.25} / 0.6892 / \textbf{0.3078} \\
         & base & 14.18 / 0.6530 / 0.5849 & 20.55 / \textbf{0.6965} / 0.3086 & 17.34 / 0.5358 / 0.8291 & 19.44 / 0.5808 / 0.5852 & \textbf{20.93} / 0.6777 / \textbf{0.3044} \\
         & cafe & 14.60 / 0.6645 / 0.4988 & \textbf{20.35} / \textbf{0.7503} / \textbf{0.2685} & 17.15 / 0.5462 / 0.7526 & 19.04 / 0.6515 / 0.4789 & 20.23 / 0.7424 / 0.2699 \\
         & canteen & 14.00 / 0.6885 / 0.5026 & 18.83 / \textbf{0.6716} / 0.3801 & 17.36 / 0.5841 / 0.7301 & 17.60 / 0.5892 / 0.5989 & \textbf{19.13} / 0.6658 / \textbf{0.3622} \\
         & center & 15.96 / 0.7035 / 0.4507 & 20.66 / 0.7020 / 0.3799 & 18.11 / 0.6290 / 0.7740 & 21.46 / 0.6756 / 0.5736 & \textbf{22.40} / \textbf{0.7477} / \textbf{0.3135} \\
         & center1 & 15.77 / \textbf{0.7330} / 0.4300 & 18.70 / 0.7082 / 0.4186 & 18.52 / 0.6697 / 0.7354 & 21.56 / 0.6930 / 0.5805 & \textbf{22.15} / 0.7221 / \textbf{0.3260} \\
         & corridor & 16.32 / \textbf{0.7468} / 0.4231 & 21.07 / 0.7329 / 0.3289 & 18.70 / 0.6680 / 0.6500 & 21.12 / 0.6764 / 0.5130 & \textbf{21.73} / 0.7335 / \textbf{0.2666} \\
         & innovation & 14.47 / 0.6423 / 0.5076 & 21.07 / \textbf{0.6902} / 0.3302 & 18.84 / 0.5711 / 0.7583 & 20.72 / 0.6279 / 0.4997 & \textbf{21.54} / 0.6824 / \textbf{0.3212} \\
         & lab & 15.33 / 0.7626 / 0.4248 & 22.77 / 0.8098 / 0.2268 & 18.92 / 0.6622 / 0.6989 & 20.54 / 0.7110 / 0.4526 & \textbf{23.15} / \textbf{0.8172} / \textbf{0.1566} \\
         & library & 16.01 / 0.6325 / 0.5107 & \textbf{22.71} / \textbf{0.6480} / 0.3662 & 18.02 / 0.5884 / 0.7884 & 21.39 / 0.5926 / 0.6001 & 22.46 / 0.6435 / \textbf{0.2964} \\
         & office & 15.64 / 0.6847 / 0.4650 & 16.47 / 0.5937 / 0.5469 & 19.04 / 0.6108 / 0.7479 & \textbf{21.62} / 0.6246 / 0.5658 & 18.94 / \textbf{0.6516} / \textbf{0.3787} \\
         \cmidrule{2-7}
         & average & 15.07 / 0.6848 / 0.4839 & 20.17 / 0.7001 / 0.3536 & 18.00 / 0.5988 / 0.7488 & 20.45 / 0.6358 / 0.5334 & \textbf{21.08} / \textbf{0.7066} / \textbf{0.3003} \\
        \midrule
        \multirow{12}{*}{100min} & bar & 14.12 / 0.6370 / 0.5174 & 18.64 / 0.6633 / 0.3487 & 16.06 / 0.5196 / 0.7395 & 19.35 / 0.6362 / 0.3194 & \textbf{19.68} / \textbf{0.7152} / \textbf{0.2250} \\
         & base & 14.73 / 0.6634 / 0.5930 & 20.83 / 0.7057 / 0.2756 & 17.40 / 0.5158 / 0.7792 & 20.05 / 0.6214 / 0.4797 & \textbf{21.33} / \textbf{0.7098} / \textbf{0.1945} \\
         & cafe & 14.96 / 0.6666 / 0.5169 & 19.03 / 0.6672 / 0.3333 & 17.30 / 0.5397 / 0.7127 & 19.56 / 0.6912 / 0.3876 & \textbf{20.43} / \textbf{0.7641} / \textbf{0.1783} \\
         & canteen & 14.20 / 0.6761 / 0.5206 & 17.51 / 0.6100 / 0.4184 & 17.47 / 0.5799 / 0.7118 & 18.36 / 0.6192 / 0.5251 & \textbf{19.06} / \textbf{0.6661} / \textbf{0.2733} \\
         & center & 15.70 / 0.6982 / 0.4644 & 20.79 / 0.6847 / 0.3846 & 18.22 / 0.6217 / 0.7426 & 22.09 / 0.6991 / 0.4954 & \textbf{22.79} / \textbf{0.7629} / \textbf{0.2094} \\
         & center1 & 15.27 / \textbf{0.7203} / 0.4557 & 19.91 / 0.7017 / 0.3990 & 18.57 / 0.6629 / 0.7072 & \textbf{22.17} / 0.7152 / 0.5039 & 18.36 / 0.6531 / \textbf{0.3337} \\
         & corridor & 16.21 / 0.7333 / 0.4537 & 18.47 / 0.6596 / 0.4131 & 19.10 / 0.6638 / 0.6212 & 21.07 / 0.6912 / 0.4632 & \textbf{21.96} / \textbf{0.7337} / \textbf{0.2106} \\
         & innovation & 14.86 / 0.6412 / 0.5162 & 20.56 / 0.6552 / 0.3483 & 19.06 / 0.5671 / 0.7170 & 21.71 / 0.6778 / 0.3931 & \textbf{21.93} / \textbf{0.7131} / \textbf{0.2089} \\
         & lab & 15.87 / 0.7538 / 0.4560 & 20.79 / 0.7280 / 0.3166 & 19.14 / 0.6589 / 0.6508 & 22.14 / 0.7600 / 0.3263 & \textbf{23.39} / \textbf{0.8258} / \textbf{0.1139} \\
         & library & 16.06 / 0.6258 / 0.5453 & 19.95 / 0.5715 / 0.4528 & 17.77 / 0.5495 / 0.7623 & \textbf{22.39} / 0.6295 / 0.4988 & 22.02 / \textbf{0.6341} / \textbf{0.2209} \\
         & office & 15.93 / \textbf{0.6783} / 0.4888 & 16.22 / 0.5867 / 0.5304 & 19.29 / 0.6059 / 0.7021 & \textbf{22.23} / 0.6485 / 0.4960 & 18.44 / 0.6439 / \textbf{0.3109} \\
         \cmidrule{2-7}
         & average & 15.26 / 0.6813 / 0.5025 & 19.34 / 0.6576 / 0.3837 & 18.12 / 0.5895 / 0.7133 & \textbf{21.18} / 0.6718 / 0.4444 & 20.85 / \textbf{0.7111} / \textbf{0.2254} \\
        \bottomrule
    \end{tabular}
    }
    \label{tab:360roam_performance_comparison}
\end{table}
\begin{table}[t]
    \centering
    \caption{Quantitative comparison of 3D reconstruction results on OmniScenes dataset (10min/100min).
    The best result for each metric is written in \textbf{bold}.
    Our method shows the best performance on almost all settings.
    }
    \resizebox{\textwidth}{!}{
    \begin{tabular}{c|c|c|c|c|c|c}
        \toprule[1.0pt]
        opt time & scene & Nerf(P) & 3DGS(P) & TensoRF & EgoNeRF & ODGS \\
        \midrule
        \multirow{8}{*}{10min} & pyebaekRoom 1 & 14.76 / 0.5940 / 0.5417 & 20.56 / 0.7347 / 0.2525 & 21.69 / 0.6800 / 0.4751 & 21.13 / 0.7055 / 0.3934 & \textbf{22.78} / \textbf{0.8038} / \textbf{0.1449} \\
         & room 1 & 15.14 / 0.7264 / 0.4518 & 22.48 / 0.8566 / 0.1985 & 19.81 / 0.8159 / 0.3020 & 21.68 / 0.7908 / 0.3358 & - / - / - \\
         & room 2 & 15.09 / 0.7156 / 0.4563 & 22.90 / 0.8277 / 0.1979 & 23.24 / 0.8042 / 0.3425 & 22.51 / 0.7837 / 0.3411 & \textbf{24.17} / \textbf{0.8303} / \textbf{0.1302} \\
         & room 3 & 16.21 / 0.7811 / 0.3790 & 25.13 / \textbf{0.8860} / 0.1554 & \textbf{26.39} / 0.8708 / 0.3022 & 22.79 / 0.8423 / 0.3454 & 24.08 / 0.8743 / \textbf{0.1384} \\
         & room 4 & 15.52 / 0.7467 / 0.4202 & 25.61 / 0.8816 / 0.1656 & 24.97 / 0.8583 / 0.2901 & 24.66 / 0.8434 / 0.3117 & \textbf{26.14} / \textbf{0.8906} / \textbf{0.1052} \\
         & room 5 & 16.75 / 0.8105 / 0.3805 & 24.43 / \textbf{0.8915} / 0.1843 & \textbf{25.52} / 0.8855 / 0.3198 & 22.87 / 0.8583 / 0.3416 & 24.49 / 0.8819 / \textbf{0.1493} \\
         & weddingHall 1 & 16.40 / 0.6783 / 0.5529 & 24.18 / 0.8326 / 0.8304 & 23.42 / 0.7680 / 0.4419 & 23.85 / 0.7741 / 0.3552 & \textbf{24.83} / \textbf{0.8347} / \textbf{0.1664} \\
         \cmidrule{2-7}
         & average & 15.79 / 0.7210 / 0.4551 & 23.80 / 0.8424 / 0.2977 & 24.21 / 0.8111 / 0.3619 & 22.97 / 0.8012 / 0.3481 & \textbf{24.42} / \textbf{0.8526} / \textbf{0.1391} \\
        \midrule
        \multirow{8}{*}{100min} & pyebaekRoom 1 & 15.99 / 0.5932 / 0.5669 & 17.71 / 0.6282 / 0.4094 & 22.57 / 0.6993 / 0.4026 & \textbf{23.69} / 0.7688 / 0.2787 & 22.91 / \textbf{0.8068} / \textbf{0.1322} \\
         & room 1 & 14.64 / 0.6604 / 0.4850 & 11.49 / 0.5864 / 0.5746 & 19.99 / 0.8226 / 0.2797 & 22.44 / 0.8083 / 0.2715 & - / - / - \\
         & room 2 & 14.66 / 0.6464 / 0.5138 & 11.68 / 0.6005 / 0.5027 & 23.58 / 0.8083 / 0.3037 & \textbf{23.53} / 0.8086 / 0.2786 & 23.13 / \textbf{0.8108} / \textbf{0.1422} \\
         & room 3 & 16.37 / 0.7657 / 0.4280 & 20.13 / 0.8066 / 0.3109 & \textbf{27.26} / 0.8793 / 0.2631 & 26.43 / 0.8789 / 0.2364 & 25.56 / \textbf{0.8847} / \textbf{0.1150} \\
         & room 4 & 16.13 / 0.7182 / 0.4724 & 19.18 / 0.7712 / 0.3390 & 26.06 / 0.8682 / 0.2495 & 26.17 / 0.8693 / 0.2336 & \textbf{26.25} / \textbf{0.8879} / \textbf{0.1025} \\
         & room 5 & 17.43 / 0.7996 / 0.4195 & 17.93 / 0.8088 / 0.3336 & 26.01 / \textbf{0.8894} / 0.2834 & \textbf{26.06} / 0.8803 / 0.2652 & 24.44 / 0.8797 / \textbf{0.1383} \\
         & weddingHall 1 & 16.64 / 0.6398 / 0.5541 & 21.85 / 0.7818 / 0.2642 & 24.04 / 0.7786 / 0.3818 & \textbf{25.00} / 0.8052 / 0.2718 & 24.77 / \textbf{0.8331} / \textbf{0.1392} \\
         \cmidrule{2-7}
         & average & 16.20 / 0.6838 / 0.4925 & 18.08 / 0.7329 / 0.3600 & 24.92 / 0.8205 / 0.3140 & \textbf{25.15} / 0.8352 / 0.2607 & 24.51 / \textbf{0.8505} / \textbf{0.1282} \\
        \bottomrule
    \end{tabular}
    }
    \label{tab:omniscenes_performance_comparison}
\end{table}
\begin{table}[t]
    \centering
    \caption{Quantitative comparison of 3D reconstruction results on 360VO dataset (10min/100min).
    The best result for each metric is written in \textbf{bold}.
    Our method shows the best performance on almost all settings.
    }
    \resizebox{\textwidth}{!}{
    \begin{tabular}{c|c|c|c|c|c|c}
        \toprule[1.0pt]
        opt time & scene & Nerf(P) & 3DGS(P) & TensoRF & EgoNeRF & ODGS \\
        \midrule
        \multirow{11}{*}{10min} & seq0 & 15.72 / 0.6574 / 0.4639 & 17.70 / 0.7101 / 0.3946 & 19.61 / 0.6975 / 0.5622 & \textbf{21.46} / 0.7229 / 0.4575 & 20.99 / \textbf{0.7750} / \textbf{0.2983} \\
         & seq1 & 15.58 / 0.5683 / 0.5255 & 24.31 / 0.7986 / 0.2639 & 20.59 / 0.6448 / 0.6457 & 23.33 / 0.6976 / 0.4372 & \textbf{26.42} / \textbf{0.8297} / \textbf{0.2069} \\
         & seq2 & 18.04 / 0.5774 / 0.5304 & 30.44 / 0.9055 / 0.1283 & 23.69 / 0.6830 / 0.5402 & 27.46 / 0.7945 / 0.3068 & \textbf{31.61} / \textbf{0.9195} / \textbf{0.0779} \\
         & seq3 & 15.81 / 0.5165 / 0.5740 & 20.50 / 0.6455 / 0.4518 & 20.53 / 0.5789 / 0.6819 & 20.10 / 0.5768 / 0.5903 & \textbf{21.72} / \textbf{0.6997} / \textbf{0.3430} \\
         & seq4 & 14.97 / 0.5853 / 0.5469 & 27.81 / 0.8646 / 0.1686 & 21.73 / 0.6673 / 0.5616 & 24.51 / 0.7395 / 0.3755 & \textbf{28.31} / \textbf{0.8917} / \textbf{0.1116} \\
         & seq5 & 15.97 / 0.6765 / 0.4242 & 19.20 / 0.7601 / 0.3380 & 19.68 / 0.7295 / 0.4952 & 21.18 / 0.7595 / 0.3971 & \textbf{21.69} / \textbf{0.8487} / \textbf{0.2034} \\
         & seq6 & 15.49 / 0.5380 / 0.5104 & \textbf{26.01} / \textbf{0.8018} / \textbf{0.2436} & 19.61 / 0.5996 / 0.5842 & 21.83 / 0.6470 / 0.4595 & 25.10 / 0.7858 / 0.2518 \\
         & seq7 & 15.46 / 0.6421 / 0.5045 & 25.42 / 0.8273 / 0.2419 & 11.13 / 0.5474 / 0.7262 & 22.72 / 0.7224 / 0.4725 & \textbf{26.36} / \textbf{0.8388} / \textbf{0.2210} \\
         & seq8 & 14.67 / 0.6730 / 0.4519 & 21.31 / 0.7900 / 0.3202 & 21.10 / 0.7409 / 0.4910 & \textbf{23.94} / 0.7814 / 0.3694 & 23.09 / \textbf{0.8316} / \textbf{0.2439} \\
         & seq9 & 15.43 / 0.7510 / 0.4168 & 15.98 / 0.7572 / 0.4194 & 20.69 / 0.7976 / 0.5313 & 19.65 / 0.7823 / 0.4803 & \textbf{21.04} / \textbf{0.8373} / \textbf{0.2972} \\
         \cmidrule{2-7}
         & average & 15.71 / 0.6186 / 0.4949 & 22.87 / 0.7861 / 0.2970 & 19.74 / 0.6543 / 0.5876 & 22.62 / 0.7224 / 0.4346 & \textbf{24.63} / \textbf{0.8245} / \textbf{0.2175} \\
        \midrule
        \multirow{11}{*}{100min} & seq0 & 17.16 / 0.6623 / 0.4865 & 18.31 / 0.7185 / 0.3880 & 19.92 / 0.6959 / 0.5528 & \textbf{22.25} / 0.7466 / 0.4081 & 21.69 / \textbf{0.8062} / \textbf{0.2214} \\
         & seq1 & 17.16 / 0.5843 / 0.5439 & 24.59 / 0.8081 / 0.2451 & 21.06 / 0.6505 / 0.6221 & 24.43 / 0.7352 / 0.3685 & \textbf{28.40} / \textbf{0.8777} / \textbf{0.0961} \\
         & seq2 & 19.25 / 0.5867 / 0.5350 & 29.99 / 0.9014 / 0.1298 & 24.44 / 0.7032 / 0.4944 & 29.22 / 0.8492 / 0.2230 & \textbf{30.58} / \textbf{0.9187} / \textbf{0.0644} \\
         & seq3 & 18.01 / 0.5398 / 0.5867 & 20.67 / 0.6464 / 0.4394 & 21.16 / 0.5850 / 0.6533 & 21.01 / 0.6101 / 0.5366 & \textbf{22.59} / \textbf{0.7632} / \textbf{0.2271} \\
         & seq4 & 17.27 / 0.6053 / 0.5461 & 25.64 / 0.8336 / 0.2285 & 22.40 / 0.6769 / 0.5307 & 26.46 / 0.7949 / 0.2737 & \textbf{30.20} / \textbf{0.9298} / \textbf{0.0564} \\
         & seq5 & 17.31 / 0.6816 / 0.4423 & 19.74 / 0.7682 / 0.3367 & 19.95 / 0.7239 / 0.4945 & 22.19 / 0.7867 / 0.3422 & \textbf{22.86} / \textbf{0.8778} / \textbf{0.1398} \\
         & seq6 & 17.88 / 0.5691 / 0.5072 & 26.36 / 0.8063 / 0.2475 & 20.26 / 0.6086 / 0.5664 & 23.93 / 0.7021 / 0.3775 & \textbf{29.07} / \textbf{0.8982} / \textbf{0.0807} \\
         & seq7 & 17.94 / 0.6729 / 0.5099 & 24.90 / 0.8025 / 0.2896 & 11.01 / 0.5349 / 0.7288 & 24.43 / 0.7599 / 0.3903 & \textbf{28.07} / \textbf{0.8833} / \textbf{0.1252} \\
         & seq8 & 17.61 / 0.6948 / 0.4818 & 21.15 / 0.7761 / 0.3391 & 21.72 / 0.7435 / 0.4811 & \textbf{25.29} / 0.8126 / 0.3132 & 24.01 / \textbf{0.8673} / \textbf{0.1800} \\
         & seq9 & 18.23 / 0.7764 / 0.4250 & 15.94 / 0.7610 / 0.4177 & 21.19 / 0.7981 / 0.5162 & 19.90 / 0.7935 / 0.4463 & \textbf{20.09} / \textbf{0.8290} / \textbf{0.2725} \\
         \cmidrule{2-7}
         & average & 17.78 / 0.6373 / 0.5064 & 22.73 / 0.7822 / 0.3061 & 20.50 / 0.6885 / 0.5531 & 23.91 / 0.7591 / 0.3679 & \textbf{26.68} / \textbf{0.8694} / \textbf{0.1264} \\
        \bottomrule
    \end{tabular}
    }
    \label{tab:360vo_performance_comparison}
\end{table}

\end{document}